\documentclass[11pt]{article}

\usepackage[a4paper, left=3cm, right=3cm, top=3cm, bottom=3cm]{geometry}%

\usepackage[T1]{fontenc}    
\usepackage{hyperref}       
\usepackage{url}            
\usepackage{booktabs}       
\usepackage{amsfonts}       
\usepackage{nicefrac}       
\usepackage{microtype}      
\usepackage{color}          
\usepackage{relsize}

\usepackage{times}
\usepackage{graphicx} 
\usepackage{subfigure}

\usepackage[round]{natbib}
\usepackage{algorithm}
\usepackage{algorithmic}
\usepackage{tikz}

\usepackage{fullpage}
\usepackage{hyperref}
\usepackage{authblk}

\usepackage{mathtools}

\usepackage{amsmath}
\usepackage{amsthm}
\usepackage{amssymb}
\usepackage{graphicx}
\usepackage{bbm}

\newtheorem{theorem}{Theorem}[section]

\theoremstyle{remark}
\newtheorem{remark}{Remark}
\newcommand{\bG}{\boldsymbol{G}}

\newcommand{\bC}{\boldsymbol{C}}

\newcommand{\bK}{\boldsymbol{K}}
\newcommand{\bI}{\boldsymbol{I_d}}

\newcommand{\bLam}{\boldsymbol{\Lambda}}
\newcommand{\bL}{\boldsymbol{L}}
\newcommand{\bR}{\boldsymbol{R}}

\newcommand{\bO}{\boldsymbol{O}}

\newcommand{\bN}{\boldsymbol{N}}

\newcommand{\bKc}{\boldsymbol{K^c}}
\newcommand{\bKcT}{\boldsymbol{K^c_T}}

\newcommand{\E}{\mathbb{E}}
\newcommand{\R}{\mathbb{R}}

\DeclarePairedDelimiter{\norm}{\lVert}{\rVert}

\begin{document}

\title{Uncovering Causality from Multivariate Hawkes Integrated Cumulants}

\author[1]{Massil Achab\footnote{massil.achab@m4x.org}}
\author[1]{Emmanuel Bacry}
\author[1]{St\'ephane Gaiffas}
\author[2]{Iacopo Mastromatteo}
\author[1,3]{Jean-Fran\c{c}ois Muzy}
\affil[1]{Centre de Math\'ematiques Appliqu\'ees, CNRS, Ecole Polytechnique, UMR 7641, 91128 Palaiseau, France}
\affil[2]{Capital Fund Management, 23 rue de l'Universit\'e, 75007 Paris, France}
\affil[3]{Laboratoire Sciences Pour l'Environnement, Universit\'e de Corse, 7 Avenue Jean Nicoli, 20250 Corte, France}

\renewcommand\Affilfont{\itshape\small}

\maketitle

\begin{abstract}
We design a new nonparametric method that allows one to estimate the matrix of integrated kernels of a multivariate Hawkes process.
This matrix not only encodes the mutual influences of each
node of the process, but also disentangles the causality relationships between them.
Our approach is the first that leads to an estimation of this matrix \emph{without any parametric modeling and estimation of the kernels themselves}.
As a consequence, it can give an estimation of causality relationships between nodes (or users), based on their activity timestamps (on a social network for instance), without knowing or estimating the shape of the activities lifetime.
For that purpose, we introduce a moment matching method that fits the second-order and the third-order integrated cumulants of the process.
A theoretical analysis allows us to prove that this new estimation technique is consistent.
Moreover, we show, on numerical experiments, that our approach is indeed very robust with respect to the shape of the kernels and
gives appealing results on the MemeTracker database and on financial order book data. \\

\end{abstract}

\section{Introduction}

In many applications, one needs to deal with data containing a very large number of irregular timestamped events that are recorded in continuous time.
These events can reflect, for instance, the activity of users on a social network, see~\cite{darpa16}, the high-frequency variations of signals in finance, see~\cite{bacry2015hawkes}, the earthquakes and aftershocks in geophysics, see~\cite{ogata1998space}, the crime activity, see~\cite{mohler2011self} or the position of genes in genomics, see~\cite{reynaud2010adaptive}.
The succession of the precise timestamps carries a great deal of information about the dynamics of the underlying systems.
In this context, multidimensional counting processes based models play a paramount role.
Within this framework, an important task is to recover the mutual influence of the nodes (i.e., the different components of the counting process), by leveraging on their timestamp patterns, see, for instance,~\cite{bacry14,lemonnier2014nonparametric,lewis2011nonparametric,zhou2013learning,gomez13, gomez15,xu2016learning}.

Consider a set of nodes $I = \{ 1, \ldots, d \}$. For each $i \in I$, we observe a set $Z^i$ of \emph{events}, where each $\tau \in Z^i$ labels the occurrence time of an event related to the activity of $i$.
The events of all nodes can be represented as a vector of counting processes $\boldsymbol{N}_t = [N_t^1 \cdots N_t^d]^\top$, where $N^i_t$ counts the number of events of node $i$ until time $t \in \R^+$, namely $N_t^i = \sum_{\tau \in Z^i} \mathbbm{1}_{\{\tau \le t\}}$.
The vector of stochastic intensities $\boldsymbol{\lambda}_t = [\lambda_t^1 \cdots \lambda_t^d]^\top$ associated with the multivariate counting process $\boldsymbol{N}_t$ is defined as
\begin{equation*}
\lambda_t^i  = \lim_{dt\rightarrow 0} \frac{\mathbb{P}(N_{t+dt}^i - N_t^i = 1| \mathcal{F}_t)}{dt}
\end{equation*}
for $i \in I$, where the filtration $\mathcal{F}_t$ encodes the information available up to time $t$.
The coordinate $\lambda_t^i$ gives the expected instantaneous rate of event occurrence at time $t$ for node $i$.
The vector  $\boldsymbol{\lambda}_t$ characterizes the distribution of $\boldsymbol{N}_t$, see~\cite{daley2007introduction}, and patterns in the events time-series can be captured by structuring these intensities.
\\

The Hawkes process introduced in~\cite{hawkes1971point} corresponds to an autoregressive structure of the intensities in order to capture self-excitation and cross-excitation of nodes, which is a phenomenon typically observed, for instance, in social networks, see for
instance~\cite{crane2008robust}.
Namely, $\boldsymbol{N}_t$ is called a \emph{Hawkes point process}
if the stochastic intensities can be written as
\begin{equation*}
	\lambda_t^i = \mu^i + \sum_{j=1}^d \int_0^t \phi^{ij}(t-t') dN_{t'}^j,
\end{equation*}
where $\mu^i \in \R^+$ is an exogenous intensity and $\phi^{ij}$ are positive, integrable and causal (with support in $\mathbb{R}_{+}$) functions called \emph{kernels} encoding the impact of an action by node~$j$ on the activity of node~$i$. Note that when all kernels are zero, the process is a simple homogeneous multivariate Poisson process.
\\

Most of the litterature uses a parametric approach for estimating the kernels. With no doubt, the most popular parametrization form is the exponential kernel $\phi^{ij}(t) = \alpha_{ij}\beta_{ij} e^{-\beta_{ij} t}$ because it definitely simplifies the inference algorithm (e.g., the complexity needed for computing the likelihood is much smaller). When $d$ is large, in order to reduce the number of parameters, some authors choose to arbitrarily share the kernel shapes across the different nodes. Thus, for instance, in \cite{yang2013mixture, zhou_zha_le_2013, gomez15}, they choose $\phi^{ij}(t) = \alpha_{ij} h(t)$ with $\alpha_{ij} \in \R^+$ quantifies the intensity of the influence of $j$ on $i$ and $h(t)$ a (normalized) function that characterizes the time-profile of this influence and that is \emph{shared} by all couples of nodes $(i,j)$ (most often, it is chosen to be either exponential $h(t) = \beta e^{-\beta t}$ or power law $h(t) = \beta t^{-(\beta + 1)}$).
Both approaches are, most of the time, highly non-realistic. On the one hand there is a priori no reason for assuming that the time-profile of the influence of a node $j$ on a node $i$ does not depend on the pair $(i,j)$. On the other hand, assuming an exponential shape or a power law shape for a kernel arbitrarily imposes an event impact that is always instantly maximal and that can only decrease with time, while in practice, there may exist a latency between an event and its maximal impact.

In order to have more flexibility on the shape of the kernels, nonparametric estimation can be considered. Expectation-Maximization algorithms can be found in~\cite{lewis2011nonparametric} (for $d=1$) or in~\cite{zhou2013learning} ($d>1$).
An alternative method is proposed in~\cite{bacry14} where the nonparametric estimation is formulated as a numerical solving of a Wiener-Hopf equation.
Another nonparametric strategy considers a decomposition of kernels on a dictionary of function $h_1, \ldots, h_K$, namely $\phi^{ij}(t) = \sum_{k=1}^K a_k^{ij} h_k(t)$, where the coefficients $a_k^{ij}$ are estimated, see~\cite{hansen_reynaud_bouret_viroirard, lemonnier2014nonparametric} and \cite{xu2016learning}, where group-lasso is used to induce a sparsity pattern on the coefficients $a_{k}^{ij}$ that is shared across $k=1, \ldots, K$.

Such methods are heavy when $d$ is large, since they rely on likelihood maximization or least squares minimization within an over-parametrized space in order to gain flexibility on the shape of the kernels.
This is problematic, since the original motivation for the use of Hawkes processes is to estimate the influence and causality of nodes, the knowledge of the full parametrization of the model being of little interest for causality purpose.
\\



Our paper solves this problem with a different and more direct approach.
Instead of trying to estimate the kernels $\phi^{ij}$, we focus on the direct estimation of their \emph{integrals}. Namely, we want to estimate the matrix $\boldsymbol{G} = [g^{ij}]$ where
\begin{equation}
\label{defG}
  	g^{ij} = \int_0^{+\infty} \phi^{ij}(u) \; du \ge 0 \; \text{ for } \; 1 \leq i, j \leq d.
\end{equation}
As it can be seen from the cluster representation of Hawkes processes (\cite{hawkes1974cluster}), this integral represents the mean total number of events of type $i$ directly triggered by an event of type $j$, and then encodes a notion of \emph{causality}.
Actually, as detailed below (see Section \ref{sec:granger}), such integral can be related to the Granger causality (\cite{granger1969}).
\\

The main idea of the method we developed in this paper is to estimate the matrix $\boldsymbol{G}$ directly using a matching cumulants (or moments) method. Apart from the mean, we shall use second and third-order cumulants which correspond respectively to centered second and third-order moments.
We first compute an estimation $\widehat{\boldsymbol{M}}$ of these centered moments $M(\boldsymbol{G})$ (they are uniquely defined by~$\boldsymbol{G}$).
Then, we look for a matrix $\boldsymbol{\widehat G}$ that minimizes the $L^2$ error $\| M(\boldsymbol{\widehat{G}}) - \widehat{\boldsymbol{M}} \|^2$. Thus the integral matrix $\boldsymbol{\widehat G}$ is directly estimated without making hardly any assumptions on the shape the involved kernels.
As it will be shown, this approach turns out to be particularly robust to the kernel shapes, which is not the case of all previous Hawkes-based approaches that aim causality recovery.
We call this method NPHC (Non Parametric Hawkes Cumulant), since our approach is of nonparametric nature.
We provide a theoretical analysis that proves the consistency of the NPHC estimator.
Our proof is based on ideas from the theory of Generalized Method of Moments (GMM) but requires an original technical trick since our setting strongly departs from the standard parametric statistics with i.i.d observations.
Note that moment and cumulant matching techniques proved particularly powerful for latent topic models, in particular Latent Dirichlet Allocation, see~\cite{podosinnikova2015rethinking}.
A small set of previous works, namely~\cite{da2014hawkes, ait2010modeling}, already used method of moments with Hawkes processes, but only in a parametric setting.
Our work is the first to consider such an approach for a nonparametric counting processes framework.
\\

The paper is organized as follows: in Section \ref{sec:nphc}, we provide the background on the integrated kernels and the integrated cumulants of the Hawkes process. We then introduce the method, investigate its complexity and explain the consistency result we prove.
In Section \ref{sec:exp}, we estimate the matrix of Hawkes kernels' integrals for various simulated datasets and for real datasets, namely the MemeTracker database and financial order book data.
We then provide in Section \ref{sec:proofs} the technical details skipped in the previous parts and the proof of our consistency result.
Section \ref{sec:conclusion} contains concluding remarks.

\section{NPHC: The Non Parametric Hawkes Cumulant method}
\label{sec:nphc}

In this Section, we provide the background on integrals of Hawkes kernels and integrals of Hawkes cumulants. We then explain how the NPHC method enables estimating $\boldsymbol{G}$.

\subsection{Branching structure and Granger causality}
\label{sec:granger}

From the definition of Hawkes process as a Poisson cluster process, see~\cite{ hk_cumul} or \cite{hawkes1974cluster}, $g^{ij}$ can be simply interpreted as the average total number of events of node~$i$ whose \emph{direct} ancestor is a given event of node~$j$ (by direct we mean that interactions mediated by any other intermediate event are not counted).
In that respect, $\boldsymbol{G}$ not only describes the mutual influences between nodes, but it also quantifies their \emph{direct causal} relationships.
Namely, introducing the counting function $N^{i \leftarrow j}_t$ that counts the number of events of $i$ whose direct ancestor is an event of $j$,
we know from~\cite{bacry2015hawkes} that
\begin{equation}
\label{eq:contrib}
\E [ d N^{i \leftarrow j}_t ] = g^{ij} \mathbb{E} [ d N^{j}_t ] = g^{ij} \Lambda^j dt,
\end{equation}
where we introduced $\Lambda^i$ as the intensity expectation, namely satisfying $\E[d N_t^i] = \Lambda^i dt$.
Note that $\Lambda^i$ does not depend on time by stationarity of $\boldsymbol N_t$, which is known to hold under the \emph{stability condition} $\| \boldsymbol G \| < 1$, where $\|\boldsymbol G \|$ stands for the spectral norm of $\boldsymbol{G}$.
In particular, this condition implies the non-singularity of $\bI - \bG$.

Since the question of a \emph{real causality} is too complex in general, most econometricians agreed on the simpler definition of Granger causality~\cite{granger1969}. Its mathematical formulation is a statistical hypothesis test: $X$ causes $Y$ \emph{in the sense of Granger causality} if forecasting future values of $Y$ is more successful while taking $X$ past values into account.
In~\cite{graphicalModelingHawkes}, it is shown that for $\boldsymbol{N}_t$ a multivariate Hawkes process, $N^j_t$ does not Granger-cause $N^i_t$ w.r.t $\boldsymbol{N}_t$ if and only if $\phi^{ij}(u) = 0$ for $u \in \mathbb{R}_{+}$. Since the kernels take positive values, the latter condition is equivalent to $\int_0^\infty \phi^{ij}(u) du = 0$.
In the following, we'll refer to \emph{learning the kernels' integrals} as \emph{uncovering causality} since each integral encodes the notion of Granger causality, and is also linked to the number of events directly caused from a node to another node, as described above at Eq.~(\ref{eq:contrib}).

\subsection{Integrated cumulants of the Hawkes process}

A general formula for the integral of the cumulants of a multivariate Hawkes process is provided in~\cite{hk_cumul}. As explained below, for the purpose of our method, we only need to consider cumulants up to the third order.
Given $1 \leq i,j,k \leq d$, the first three integrated cumulants of the Hawkes process can be defined as follows thanks to stationarity:
\begin{align}
	\label{eq:cumul1density}
  \Lambda^i dt &= \mathbb{E}(dN_t^i) \\
    \label{eq:cumul2density}
    C^{ij} dt &= \int_{\tau \mathsmaller{\in} \mathbb{R}} \! \Big(
    \mathbb{E}(dN_{t}^{i} dN_{t+\tau}^{j}) - \mathbb{E}(dN_{t}^{i}) \mathbb{E}(dN_{t+\tau}^{j}) \Big) \\
    \begin{split}
    	  K^{ijk} dt &= \int \! \! \int_{\tau,\tau' \mathsmaller{\in} \mathbb{R}^2} \! \Big(
    \mathbb{E}(dN^i_t dN^j_{t+\tau} dN^k_{t+\tau'}) + 2 \mathbb{E}(dN^i_t) \mathbb{E}(dN^j_{t+\tau}) \mathbb{E}(dN^k_{t+\tau'}) \\
    & \quad \quad \quad - \mathbb{E}(dN^i_t dN^j_{t+\tau}) \mathbb{E}(dN^k_{t+\tau'}) - \mathbb{E}(dN^i_t dN^k_{t+\tau'}) \mathbb{E}(dN^j_{t+\tau})  - \mathbb{E}(dN^j_{t+\tau} dN^k_{t+\tau'}) \mathbb{E}(dN^i_t) \Big),
    \end{split}
    \label{eq:cumul3density}
\end{align}
where Eq.~(\ref{eq:cumul1density}) is the mean intensity of the Hawkes process, the second-order cumulant~(\ref{eq:cumul2density}) refers to the integrated covariance density matrix and the third-order cumulant~(\ref{eq:cumul3density}) measures the skewness of~$\boldsymbol{N}_t$.
Using the martingale representation from~\cite{bacry14} or the Poisson cluster process representation from~\cite{hk_cumul}, one can obtain an explicit relationship between these integrated cumulants and the matrix
$\boldsymbol{G}$. If one sets
\begin{equation}
\label{defR}
\bR = (\bI - \bG)^{-1},
\end{equation}
straightforward computations (see Section~\ref{sec:proofs})
lead to the following identities:
\begin{align}
\label{eq:relation0}
\Lambda^i  &=  \sum_{m=1}^d R^{im} \mu^m  \\
\label{eq:relation1}
C^{ij} &=  \sum_{m=1}^d \Lambda^m R^{im} R^{jm} \\
\label{eq:relation2}
K^{ijk}  &=  \sum_{m=1}^d ( R^{im}R^{jm}C^{km} + R^{im}C^{jm}R^{km} + C^{im}R^{jm}R^{km} - 2 \Lambda^m R^{im}R^{jm}R^{km}).
\end{align}
Equations~(\ref{eq:relation1}) and (\ref{eq:relation2}) are proved in Section~\ref{sec:proofs}.
Our strategy is to use a convenient subset of Eqs.~(\ref{eq:cumul1density}),~(\ref{eq:cumul2density}) and~(\ref{eq:cumul3density}) to define ${\boldsymbol{M}}$, while we use Eqs.~(\ref{eq:relation0}),~(\ref{eq:relation1}) and~(\ref{eq:relation2}) in order to construct the operator that maps a candidate matrix $\boldsymbol R$ to the corresponding cumulants $M(\boldsymbol R)$.
By looking for $\widehat{\boldsymbol R}$ that minimizes $\boldsymbol R \mapsto \| M(\boldsymbol{R}) - \widehat{\boldsymbol{M}} \|^2$, we obtain, as illustrated below, good recovery of the ground truth matrix $\bG$ using Equation~\eqref{defR}.

The simplest case $d=1$ has been considered in~\cite{bouchaud14}, where it is shown that one can choose $M = \{C^{11}\}$ in order to compute the kernel integral. Eq.~(\ref{eq:relation1}) then reduces to a simple second-order equation that has a unique solution in  $\boldsymbol{R}$ (and consequently a unique $\boldsymbol{G}$) that accounts for the stability condition ($ \| \boldsymbol{G} \| <1$).

Unfortunately, for $d>1$, the choice $M = \{C^{ij}\}_{1\leq i\leq j \leq d}$ is not sufficient to uniquely determine the kernels integrals.
In fact, the integrated covariance matrix provides $d(d+1)/2$ independent coefficients, while $d^2$ parameters are needed.
It is straightforward to show that the remaining $d(d-1)/2$ conditions can be encoded in an orthogonal matrix $\boldsymbol O$, reflecting the fact that Eq.~(\ref{eq:relation1}) is invariant under the change $\bR \to \boldsymbol O \bR$, so that the system is under-determined.

Our approach relies on using the third order cumulant tensor $\bK = [K^{ijk}]$ which contains $(d^3+3d^2+2d)/6 > d^2$ independent coefficients that are sufficient to uniquely fix the matrix $\bG$.
This can be justified intuitively as follows: while the integrated covariance only contains symmetric information, and is thus unable to provide causal information, \emph{the skewness given by the third order cumulant in the estimation procedure can break the symmetry between past and future so as to uniquely fix $\bG$}.
Thus, our algorithm consists of selecting  $d^2$ third-order cumulant components, namely
$M= \{ K^{iij}\}_{1\leq i,j \leq d}$.
In particular, we define the estimator of $\bR$ as $\widehat \bR \in \textrm{argmin}_{\bR} \mathcal{L}(\bR)$, where
\begin{align}
	\label{eq:nphc_loss}
  &\mathcal{L}(\bR) = (1 - \kappa) \|\bKc(\bR) - \widehat{\bKc}\|_2^2 + \kappa \| \bC (\bR) - \widehat{\bC} \|_2^2,
\end{align}
where $\| \cdot \|_2$ stands for the Frobenius norm,
$\bKc = \{ K^{iij} \}_{1\leq i,j\leq d}$ is the matrix obtained by the contraction of the tensor $\bK$ to $d^2$ indices, $\bC$ is the covariance matrix, while $\widehat{\bKc}$ and $\widehat{\bC}$ are their respective estimators, see Equations~\eqref{eq:estimator2}, \eqref{eq:estimator3} below.
It is noteworthy that the above mean square error approach can be seen as a peculiar Generalized Method of Moments (GMM), see~\cite{hall}.
This framework allows us to determine the optimal weighting matrix involved in the loss function.
However, this approach is unusable in practice, since the associated complexity is too high. Indeed,
since we have $d^2$ parameters, this matrix has $d^4$ coefficients and GMM calls for computing its inverse leading to a $O(d^6)$ complexity.
In this work, we use the coefficient $\kappa$ to scale the two terms, as
\begin{equation*}
\kappa = \frac{\|\widehat{\bKc}\|^2_2}{\|\widehat{\bKc}\|^2_2 +
  \|\widehat{\bC}\|^2_2},
\end{equation*}
see Section~\ref{sec:kappa} for an explanation about the link between $\kappa$ and the weighting matrix.
Finally, the estimator of $\bG$ is straightforwardly obtained as
\begin{equation*}
	\widehat{\bG} = \bI - \widehat{\bR}^{-1},
\end{equation*}
from the inversion of Eq.~\eqref{defR}.
Let us mention an important point: the matrix inversion in the previous formula is not the bottleneck of the algorithm. Indeed, its has a complexity $O(d^3)$ that is cheap compared to the computation of the cumulants when $n = \max_i |Z^i| \gg d$, which is the typical scaling satisfied in applications.
Solving the considered problem on a larger scale, say $d \gg 10^3$, is an open question, even with state-of-the-art parametric and nonparametric approaches, see for instance~\cite{zhou2013learning, xu2016learning, zhou_zha_le_2013, bacry14}, where the number of components $d$ in experiments is always around $100$ or smaller.
Note that, actually, our approach leads to a \emph{much faster} algorithm than the considered state-of-the-art baselines, see Tables~1--4 from Section~\ref{sec:exp} below.

\subsection{Estimation of the integrated cumulants} 


In this section we present explicit formulas to estimate the three moment-based quantities listed in the previous section, namely, $\bLam$, $\bC$ and $\bK$.
We first assume there exists $H>0$ such that the truncation from $(-\infty,+\infty)$ to $[-H,H]$ of the domain of integration of the quantities appearing in Eqs.~(\ref{eq:cumul2density}) and~(\ref{eq:cumul3density}), introduces only a small error.
In practice, this amounts to neglecting border effects in the covariance density and in the skewness density that is a good approximation if the support of the kernel $\phi^{ij}(t)$ is smaller than $H$ and the spectral norm $\|\bG\|$ satisfies $\|\bG\|<1$.

\noindent
In this case, given a realization of a stationary Hawkes process $\{ \boldsymbol N_t : t \in [0, T] \}$, as shown in Section~\ref{sec:proofs}, we can write the estimators of the first three cumulants~(\ref{eq:cumul1density}),~(\ref{eq:cumul2density}) and~(\ref{eq:cumul3density}) as
\begin{align}
    \label{eq:estimator1}
    \widehat \Lambda^i &= \frac{1}{T} \sum_{\tau \in Z^i} 1 = \frac{N^i_T}{T} \, \\
    \label{eq:estimator2}
    \widehat{C}^{ij} &= \frac{1}{T} \sum_{\tau \in Z^i} \left( N^j_{\tau+H} - N^j_{\tau-H} - 2 H \widehat{\Lambda}^j \right) \, \\
    \begin{split}
    \widehat{K}^{ijk} &= \frac{1}{T} \sum_{\tau \in Z^i} \left( N^j_{\tau+H} - N^j_{\tau-H} - 2 H \widehat{\Lambda}^j \right)  \cdot\left( N^k_{\tau+H} - N^k_{\tau-H} - 2 H \widehat{\Lambda}^k \right) \\
    & \quad - \frac{\widehat \Lambda^i}{T} \sum_{\tau \in Z^j} \sum_{\tau' \in Z^k} (2 H - | \tau'-\tau|)^{+} + 4 H^2 \widehat{\Lambda}^i \widehat{\Lambda}^j \widehat{\Lambda}^k.
    \end{split}
    \label{eq:estimator3}
\end{align}

Let us mention the following facts.
\begin{description}
    \item[Bias.] While the first cumulant $\hat \Lambda^i$ is an unbiased estimator of $\Lambda^i$, the other estimators $\widehat{C}^{ij}$ and $\widehat{K}^{ijk}$ introduce a bias. However, as we will show, in practice this bias is small and hardly affects numerical estimations (see Section \ref{sec:exp}). This is confirmed by our theoretical analysis, which proves that if $H$ does not grow too fast compared to $T$, then these estimated cumulants are consistent estimators of the theoretical cumulants (see Section~\ref{subsec:theory}).
    \item[Complexity.] The computations of all the estimators of the first, second and third-order cumulants have complexity respectively $O(nd)$,~$O(nd^2)$ and $O(nd^3)$, where $n = \max_i |Z^i|$.
    However, our algorithm requires a lot less than that: it computes only $d^2$ third-order terms, of the form $\widehat{K}^{iij}$, leaving us with only $O(nd^2)$ operations to perform.
    \item[Symmetry.] While the values of $\Lambda^i, C^{ij}$ and $K^{ijk}$ are symmetric under permutation of the indices, their estimators are generally not symmetric.
    We have thus chosen to symmetrize the estimators by averaging their values over permutations of the indices.
    Worst case is for the estimator of $\bKc$, which involves only an extra factor of 2 in the complexity.
\end{description}

\subsection{The NPHC algorithm} 


The objective to minimize in Equation~\eqref{eq:nphc_loss} is non-convex.
More precisely, the loss function is a polynomial of $\bR$ of degree 6.
However, the expectations of cumulants $\bLam$ and $\bC$ defined in Eq.~(\ref{eq:cumul2density}) and~(\ref{eq:cumul3density}) that appear in the definition of $\mathcal{L}(\bR)$ are unknown and should be replaced with $\widehat{\bLam}$ and $\widehat{\bC}$.
We denote $\widetilde{\mathcal{L}}(\bR)$ the objective function, where the expectations of cumulants $\Lambda^i$ and $C^{ij}$ have been replaced with their estimators in the right-hand side of Eqs.~(\ref{eq:relation1}) and ~(\ref{eq:relation2}):
\begin{align}
\widetilde{\mathcal{L}}(\bR) &= (1 - \kappa) \| \bR^{\odot 2} \widehat{\bC}^\top + 2 [\bR \odot (\widehat{\bC} - \bR \widehat{\bL})] \bR^\top - \widehat{\bKc} \|_2^2 + \kappa \| \bR \widehat{\bL} \bR^\top - \widehat{\bC}\|_2^2
\label{eq:nphc_effective_loss}
\end{align}
As explained in~\cite{choromanska2015loss}, the loss function of a typical multilayer neural network with simple nonlinearities can be expressed as a polynomial function of the weights in the network, whose degree is the number of layers.
Since the loss function of NPHC writes as a polynomial of degree $6$, we expect good results using optimization methods designed to train deep multilayer neural networks.
We used the AdaGrad from~\cite{duchi2011adaptive}, a variant of the Stochastic Gradient Descent with adaptive learning rates.
AdaGrad scales the learning rates coordinate-wise using the online variance of the previous gradients, in order to incorporate second-order information during training.
The NPHC method is summarized schematically in Algorithm~\ref{algo:mf}.

 \begin{algorithm}
 \caption{Non Parametric Hawkes Cumulant method}
 \label{algo:mf}
 \begin{algorithmic}[1]
 \renewcommand{\algorithmicrequire}{\textbf{Input:}}
 \renewcommand{\algorithmicensure}{\textbf{Output:}}
 \REQUIRE $\boldsymbol{N}_t$
 \ENSURE  $\widehat{\bG}$
  \STATE Estimate $\widehat{\Lambda}^i$, $\widehat{C}^{ij}, \widehat{K}^{iij}$ from Eqs.~(\ref{eq:estimator1}, \ref{eq:estimator2}, \ref{eq:estimator3})
  \STATE Design $\widetilde{\mathcal{L}}(\bR)$ using the computed estimators.
  \STATE Minimize numerically $\widetilde{\mathcal{L}}(\bR)$ so as to obtain $\widehat{\bR}$
  \STATE Return $\widehat{\bG} = \bI - \widehat{\bR}^{-1}$.
 \end{algorithmic}
 \end{algorithm}

Our problem being non-convex, the choice of the starting point has a major effect on the convergence. Here, the key is to notice that the matrices $\bR$ that match Equation~(\ref{eq:relation1}) writes $\bC^{1/2} \bO \bL^{-1/2}$, with $\bL = \mbox{diag}(\bLam)$ and $\bO$ an orthogonal matrix.
Our starting point is then simply chosen by setting  $\bO = \bI$ in the previous formula, leading to nice convergence results.
Even though our main concern is to retrieve the matrix $\bG$, let us notice we can also obtain an estimation of the baseline intensities' from Eq.~(\ref{eq:cumul1density}), which leads to
 $\widehat{\boldsymbol{\mu}} = \widehat{\bR}^{-1} \widehat{\boldsymbol{\Lambda}}$.
An efficient implementation of this algorithm with TensorFlow, see \cite{abadi2016tensorflow}, is available on GitHub: \url{https://github.com/achab/nphc}.

\subsection{Complexity of the algorithm}
\label{subsec:complexity}

Compared with existing state-of-the-art methods to estimate the kernel functions,
e.g., the ordinary differential equations-based (ODE) algorithm in~\cite{zhou2013learning}, the Granger Causality-based algorithm in \cite{xu2016learning},
the ADM4 algorithm in \cite{zhou_zha_le_2013}, and the Wiener-Hopf-based algorithm in \cite{bacry14}, our method has a very competitive complexity.
This can be understood by the fact that those methods estimate the kernel functions,
while in NPHC we only estimate their integrals.
The ODE-based algorithm is an EM algorithm that parametrizes the kernel function with $M$ basis functions, each being discretized to $L$ points.
The basis functions are updated after solving $M$ Euler-Lagrange equations.
If $n$ denotes the maximum number of events per component (i.e. $n = \max_{1\leq i \leq d} |Z^i|$) then the complexity of one iteration of the algorithm is $O(M n^3 d^2 + M L (n d + n^2))$. The Granger Causality-based algorithm is similar to the previous one, without the update of the basis functions, that are Gaussian kernels.
The complexity per iteration is $O(M n^3 d^2)$.
The algorithm ADM4 is similar to the two algorithms above, as EM algorithm as well, with only one exponential kernel as basis function.
The complexity per iteration is then $O(n^3 d^2)$.
The Wiener-Hopf-based algorithm is not iterative, on the contrary to the previous ones.
It first computes the empirical conditional laws on many points, and then invert the Wiener-Hopf system, leading to a $O(n d^2 L + d^4 L^3)$ computation.
Similarly, our method first computes the integrated cumulants,
then minimize the objective function with $N_{\text{iter}}$ iterations, and invert the resulting matrix $\widehat{\bR}$ to obtain $\widehat{\bG}$.
In the end, the complexity of the NPHC method is $O(n d^2 + N_{\text{iter}} d^3)$.
According to this analysis, summarized in Table~\ref{table:complexity} below, one can
see that in the regime $n \gg d$, the NPHC method outperforms all the other ones.
\begin{table}[h]
\small
  \caption{Complexity of state-of-the-art methods. NPHC's complexity is very low , especially in the regime $n \gg d$.}
  \label{table:complexity}
  \centering
  \begin{tabular}{llllll}
    \toprule
    Method  & Total complexity \\
    \midrule
    ODE \cite{zhou2013learning} &$O(N_{\text{iter}} M (n^3 d^2 +  L (n d + n^2)))$  \\
    GC \cite{xu2016learning} &$O(N_{\text{iter}} M n^3 d^2)$  \\
    ADM4 \cite{zhou_zha_le_2013} &$O(N_{\text{iter}} n^3 d^2)$   \\
    WH \cite{bacry14} &$O(n d^2 L + d^4 L^3)$  \\
    NPHC  &$O(n d^2 + N_{\text{iter}} d^3)$  \\
    \bottomrule
  \end{tabular}
\end{table}

\subsection{Theoretical guarantee: consistency} 
\label{subsec:theory}

The NPHC method can be phrased using the framework of the Generalized Method of
Moments (GMM).
GMM is a generic method for estimating parameters in statistical models.
In order to apply GMM, we have to find a vector-valued function $g(X, \theta)$ of the data,
where $X$ is distributed with respect to a distribution $\mathbb P_{\theta_0}$,
which satisfies the \emph{moment condition}: $\mathbb{E}[g(X,\theta)]=0$ if and only if $\theta=\theta_0$, where $\theta_0$ is the ``ground truth'' value of the parameter.
Based on i.i.d. observed copies $x_1, \ldots, x_n$ of $X$, the GMM method minimizes the norm of the empirical mean over $n$ samples,
 $\|\frac{1}{n} \sum_{i=1}^n g(x_i, \theta)\|$, as a function of $\theta$, to obtain an estimate of $\theta_0$.

In the theoretical analysis of NPHC, we use ideas from the consistency proof of the GMM, but the proof actually relies on very different arguments.
Indeed, the integrated cumulants estimators used in NPHC are not unbiased, as the theory of GMM requires, but asymptotically unbiased.
Moreover, the setting considered here, where data consists of a single realization $\{ \bN_t \}$ of a Hawkes process strongly departs from the standard i.i.d setting.
Our approach is therefore based on the GMM idea but the proof is actually not using the theory of GMM.

In the following, we
use the subscript $T$ to refer to quantities that only depend on the process $(\boldsymbol{N_t})$ in the interval $[0,T]$ (e.g.,
the truncation term $H_T$, the estimated integrated covariance $\widehat{\bC}_T$ or the estimated kernel norm matrix $\widehat{\bG}_T$).
In the next equation, $\odot$ stands for the Hadamard product and $\odot 2$ stands for the entrywise square of a matrix.
We denote $\bG_0 = \bI - \bR_0^{-1}$ the true value of $\bG$, and the
 $\mathbb{R}^{2d\times d}$ valued vector functions
\begin{align*}
  g_0 (\bR) &= \begin{bmatrix}
     \bC - \bR \bL \bR^\top \\
     \bKc - \bR^{\odot 2} \bC^\top - 2 [\bR \odot (\bC - \bR \bL)] \bR^\top
  \end{bmatrix} \\
  \widehat{g}_T (\bR) &= \begin{bmatrix}
\widehat{\bC}_T - \bR \widehat{\bL}_T \bR^\top \\
\widehat{\bKcT} - \bR^{\odot 2} \widehat{\bC}_T^\top - 2 [\bR \odot (\widehat{\bC}_T - \bR \widehat{\bL}_T)] \bR^\top \; .
\end{bmatrix}
\end{align*}
Using these notations, $\widetilde{\mathcal{L}}_T(\bR)$ can be seen as the weighted squared Frobenius norm of $\widehat{g}_T(\bR)$. Moreover, when $T \rightarrow +\infty$,
one has $\widehat{g}_T(\bR) \overset{\mathbb{P}}{\rightarrow} g_0(\bR)$ under the conditions of the following theorem, where $\overset{\mathbb{P}}{\rightarrow}$ stands for convergence in probability.
\begin{theorem}[Consistency of NPHC]
Suppose that $(\boldsymbol{N_t})$ is observed on $\mathbb{R}^{+}$ and assume that
\begin{enumerate}
	\item $g_0 (\bR) = 0$ if and only if $\bR=\bR_0$\textup;
	\item $\bR \in \Theta$, where $\Theta$ is a compact set\textup;
	\item the spectral radius of the kernel norm matrix satisfies $\|\bG_0\| < 1$\textup;
\item $H_T \rightarrow \infty$ and $H_T^2 / T \rightarrow 0$.
\end{enumerate}
Then
\begin{equation*}
	\widehat{\bG}_T = \bI - \left( \arg \min_{\bR \in \Theta} \widetilde{\mathcal{L}}_T (\bR) \right)^{-1}
	 \overset{\mathbb{P}}{\rightarrow} \bG_0.
\end{equation*}
\end{theorem}
The proof of the Theorem is given in Section~\ref{sec:proof_theorem} below.
Assumption~3 is mandatory for stability of the Hawkes process, and Assumptions~3 and~4 are sufficient to prove that the estimators of the integrated cumulants defined in
Equations~\eqref{eq:estimator1}, \eqref{eq:estimator2} and \eqref{eq:estimator3} are asymptotically consistent.
Assumption~2 is a very mild standard technical assumption allowing to prove consistency for estimators based on moments.
Assumption~1 is a standard asymptotic moment condition, that allows to identity parameters from the integrated cumulants.

\section{Numerical Experiments}
\label{sec:exp}

In this Section, we provide a comparison of NPHC with the state-of-the art, on simulated datasets with different kernel shapes, the MemeTracker dataset (social networks) and the order book dynamics dataset (finance).

\paragraph*{Simulated datasets.}

We simulated several datasets with Ogata's Thinning algorithm~\cite{ogata1981lewis} using the open-source library \texttt{tick}\footnote{\url{https://github.com/X-DataInitiative/tick}}, each corresponding to a shape of kernel:
rectangular, exponential or power law kernel, see Figure~\ref{fig:kernels} below.

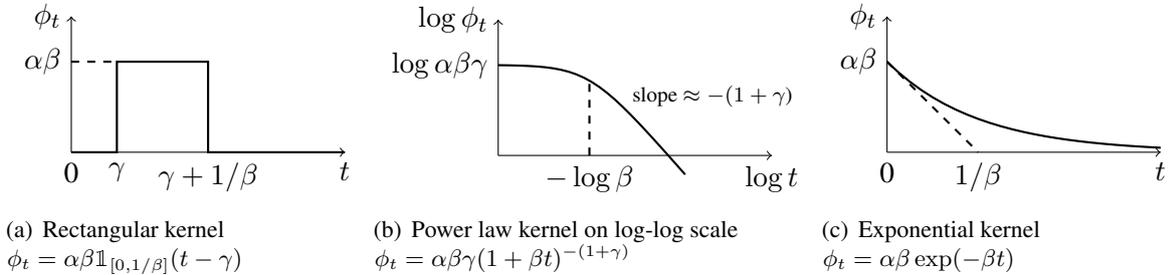
\begin{figure}[htbp]
\subfigure[Rectangular kernel
\newline $\phi_t = \alpha \beta \mathbbm{1}_{[0, 1 / \beta]}(t - \gamma) $]
{
\begin{tikzpicture}[scale=0.6]
\draw[->] (0,0) -- (6,0) node[anchor=north] {$t$};
\draw[->] (0,0) -- (0,3) node[anchor=east] {$\phi_t$};
\draw	(0,0) node[anchor=north] {0}
      (1,0) node[anchor=north] {$\gamma$}
		  (3,0) node[anchor=north] {$\gamma + 1/\beta$}
      (0,2) node[anchor=east] {$\alpha \beta$};
\draw[thick] (0,0) -- (1,0) -- (1,2) -- (3,2) -- (3,0) -- (6,0);
\draw[thick,dashed] (0,2) -- (1,2);
\end{tikzpicture}
}
\subfigure[Power law kernel on log-log scale
\newline $\phi_t = \alpha \beta \gamma (1 + \beta t)^{-(1 + \gamma)}$]
{
\begin{tikzpicture}[scale=0.6]
\draw[->] (0,0) -- (6,0) node[anchor=north] {$\log t$};
\draw[->] (0,0) -- (0,3) node[anchor=east] {$\log \phi_t$};
\draw (2,0) node[anchor=north] {$- \log \beta$}
		  (0,2) node[anchor=east] {$\log \alpha \beta \gamma$};
\draw (4.7,1.3) node{{\scriptsize slope $\approx -(1+\gamma$)}};
\draw[thick] [domain=0:4.1] plot(\x, {2-0.5*ln(1+10^(\x-2))});
\draw[thick,dashed] (2,0) -- (2,1.6);
\end{tikzpicture}
}
\subfigure[Exponential kernel \newline $\phi_t = \alpha \beta \exp (- \beta t)$]
{
\centering
\begin{tikzpicture}[scale=0.6]
\draw[->] (0,0) -- (6,0) node[anchor=north] {$t$};
\draw[->] (0,0) -- (0,3) node[anchor=east] {$\phi_t$};
\draw	(0,0) node[anchor=north] {0}
		(2,0) node[anchor=north] {$1/\beta$}
		(0,2) node[anchor=east] {$\alpha \beta$};
\draw[thick] [domain=0:6] plot(\x,{2 * exp(-0.5*\x)});
\draw[thick,dashed] (0,2) -- (2,0);
\end{tikzpicture}
}
\caption{The three different kernels used to simulate the datasets.}
\label{fig:kernels}
\end{figure}

\begin{figure}[ht]
\label{res:nphc_exp100}
\centering
\includegraphics[width=.30\textwidth]{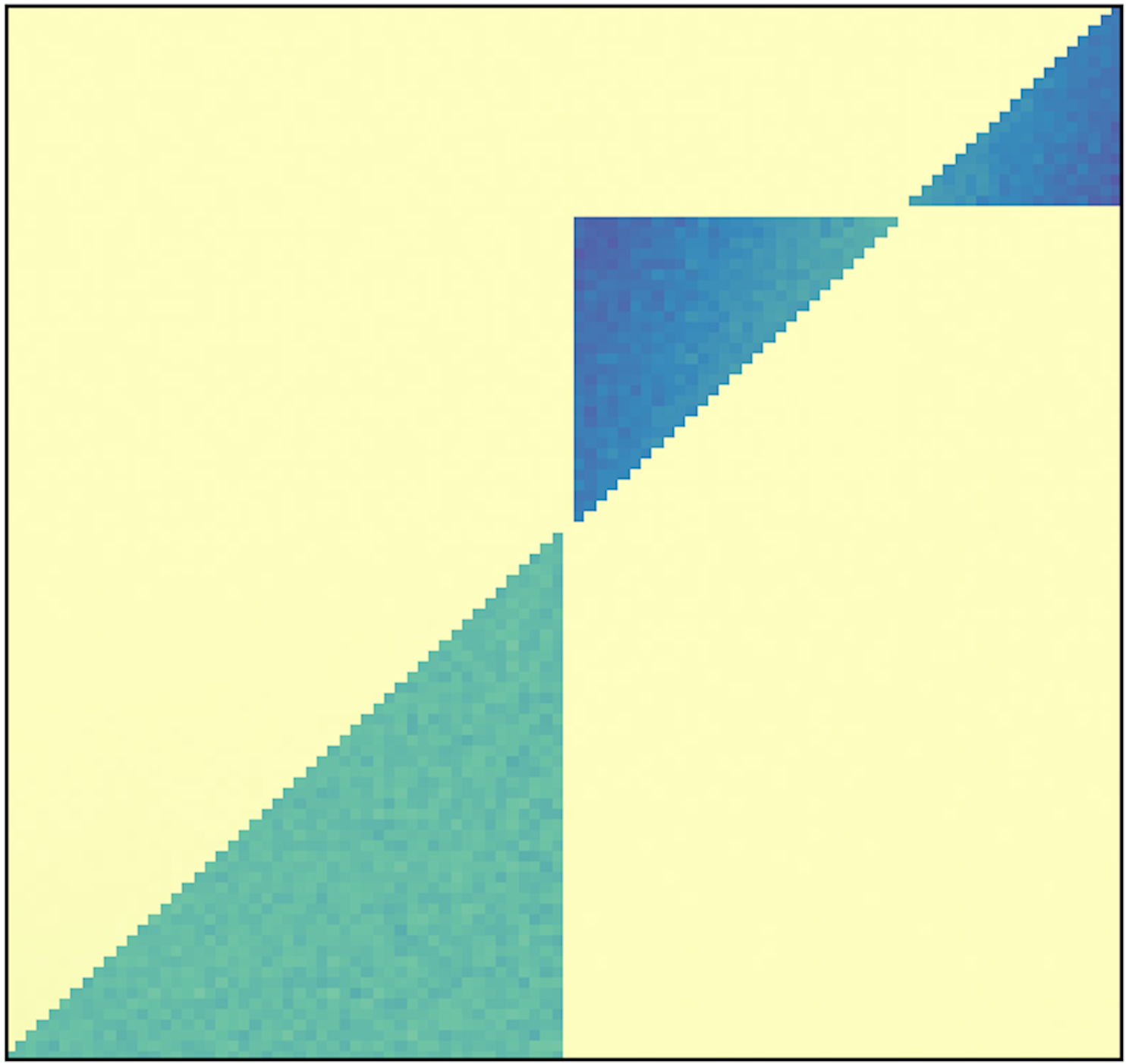}
\includegraphics[width=.30\textwidth]{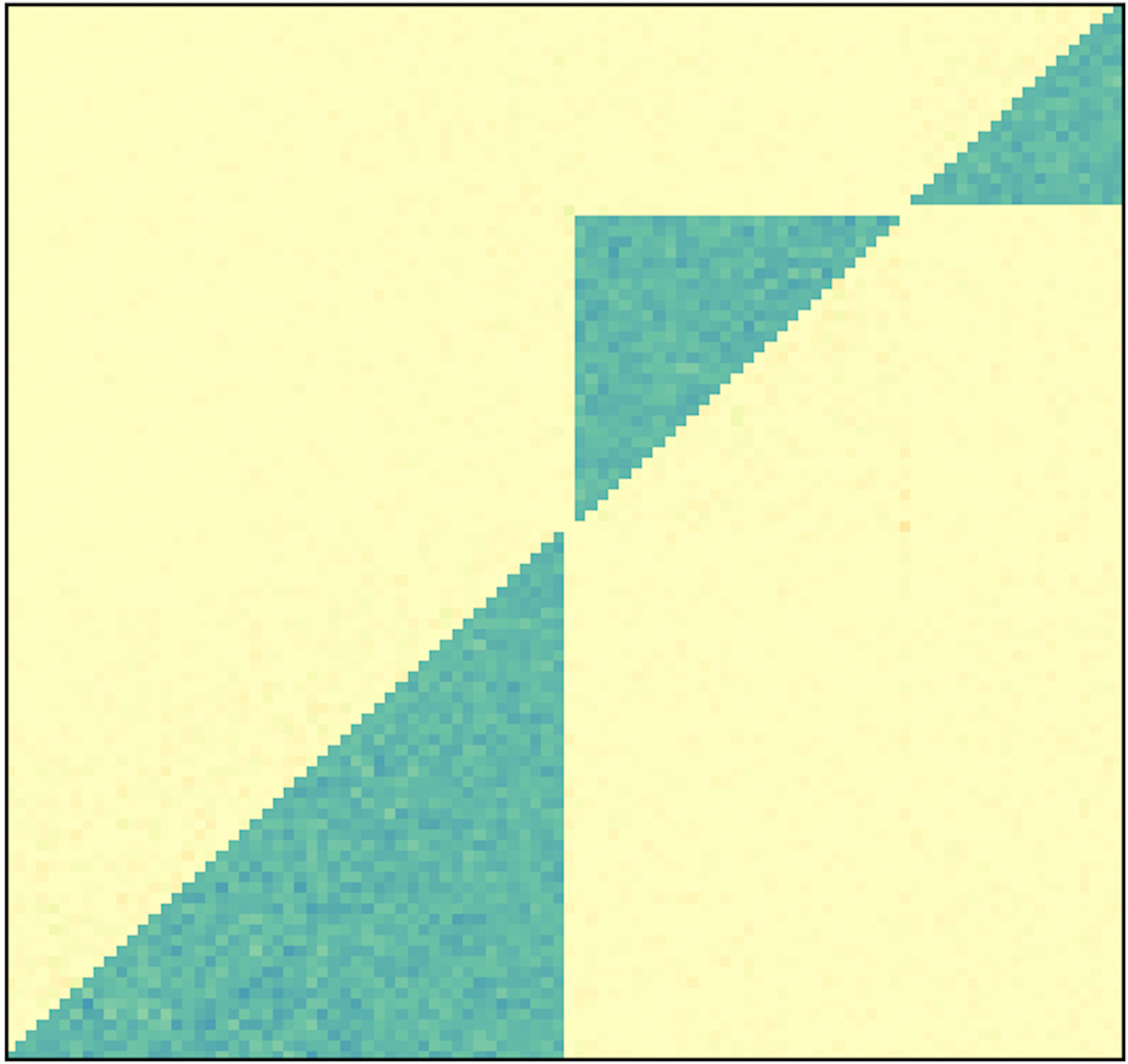}
\includegraphics[width=.30\textwidth]{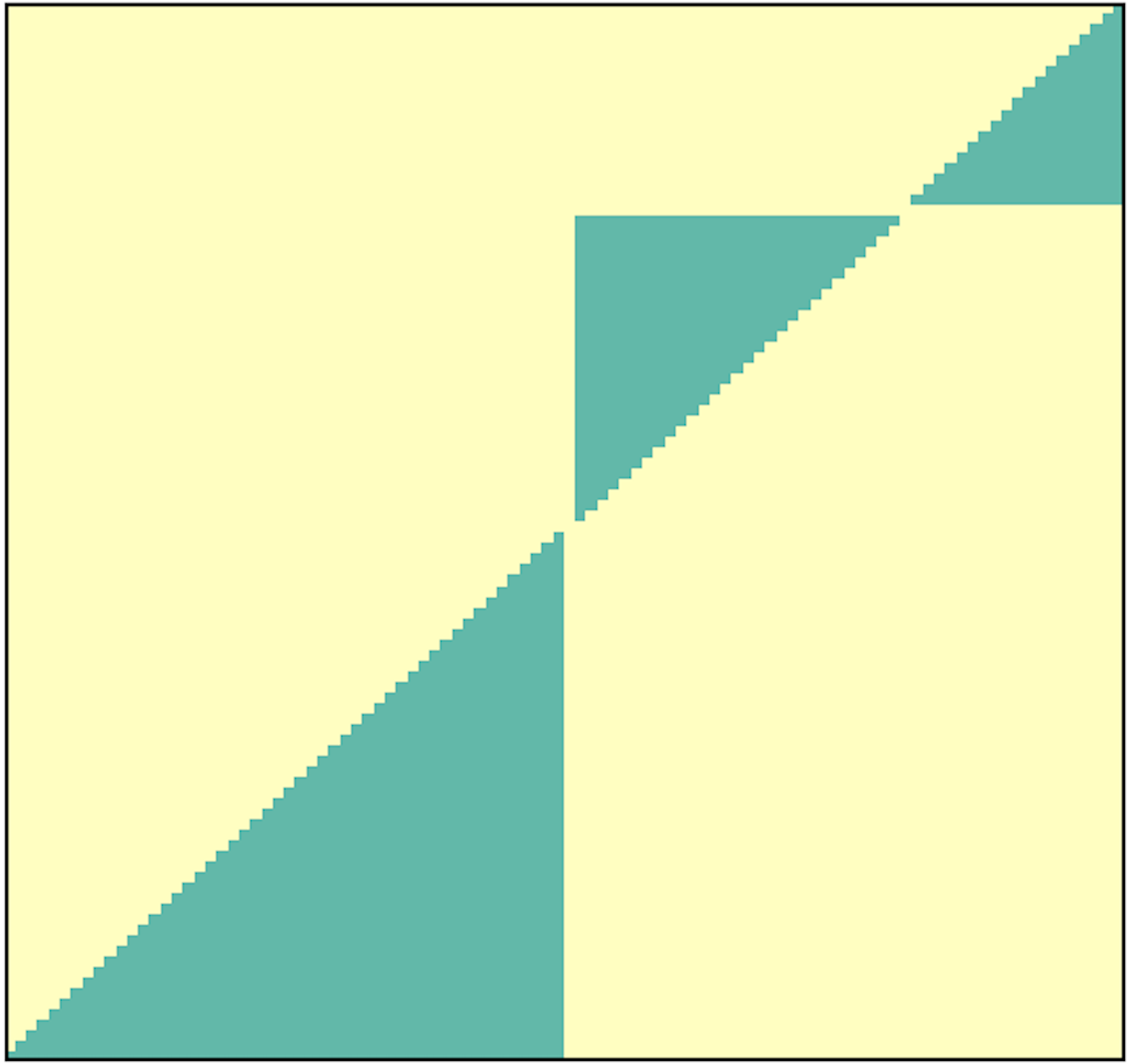}
\includegraphics[width=.06\textwidth]{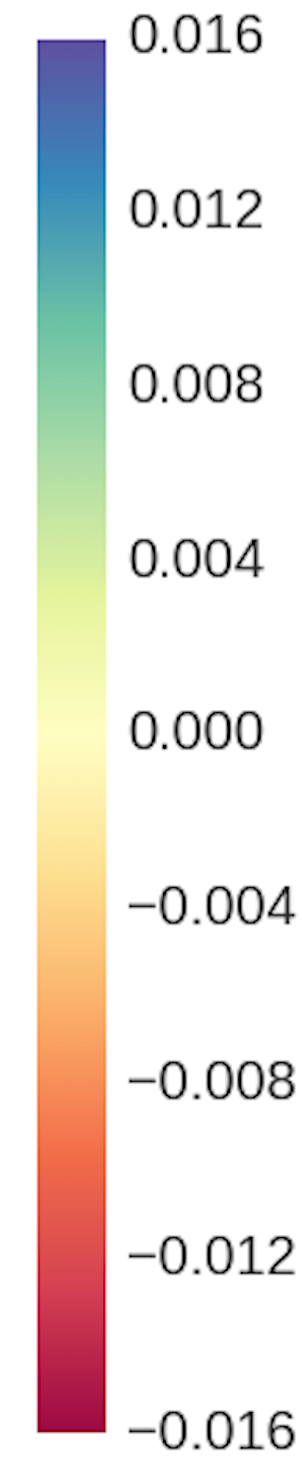}
\caption{On Exp100 dataset, estimated $\widehat{\bG}$ with ADM4 (left), with NPHC (middle) and the ground-truth matrix $\bG$ (right). Both ADM4 and NPHC estimates recover the three blocks. However, ADM4 overestimates the integrals on two of the three blocks, while NPHC gives the same value on each blocks.}
\end{figure}

The integral of each kernel on its support equals $\alpha$, $1/\beta$ can be regarded as a characteristic time-scale and $\gamma$ is the scaling exponent for the power law distribution and a delay parameter for the rectangular one.
We consider a non-symmetric block-matrix $\bG$ to show that our method can effectively uncover causality between the nodes, see Figure~\ref{res:nphc_exp100}.
The matrix $\bG$ has constant entries $\alpha$ on the three blocks - $\alpha = g^{ij} = 1 / 6$ for dimension 10 and $\alpha = g^{ij} = 1 / 10$ for dimension 100 -, and zero outside.
The two other parameters' values are the same for dimensions 10 and 100.
The parameter $\gamma$ is set to $1/2$ on the three blocks as well, but we set three very different $\beta_0$, $\beta_1$ and $\beta_2$ from one block to the other, with ratio $\beta_{i+1} / \beta_i = 10$ and $\beta_0 = 0.1$.
The number of events is roughly equal to $10^5$ on average over the nodes.
We ran the algorithm on three simulated datasets: a 10-dimensional process with rectangular kernels named Rect10, a 10-dimensional process with power law kernels named PLaw10 and a 100-dimensional process with exponential kernels named Exp100.

\paragraph{MemeTracker dataset.}

We use events of the most active sites from the MemeTracker dataset\footnote{\url{https://www.memetracker.org/data.html}}.
This dataset contains the publication times of articles in many websites/blogs from August 2008 to April 2009, and hyperlinks between posts.
We extract the top 100 media sites with the largest number of documents, with about 7 million of events.
We use the links to trace the flow of information and establish an estimated ground truth for the matrix $\bG$.
Indeed, when an hyperlink $j$ appears in a post in website $i$, the link $j$ can be regarded as a direct ancestor of the event. Then, Eq.~\eqref{eq:contrib} shows $g^{ij}$ can be estimated by $N^{i\leftarrow j}_T/ N^{j}_T = \# \{ \mbox{links } j \rightarrow i \}/ N^{j}_T$.

\paragraph{Order book dynamics.}

We apply our method to financial data, in order to understand
the self and cross-influencing dynamics of all event types in an order book.
An order book is a list of buy and sell orders for a specific financial instrument,
the list being updated in real-time throughout the day.
This model has first been introduced in \cite{2014jaisson}, and models the order book
via the following 8-dimensional point process:
$N_t = (P^{(a)}_t, P^{(b)}_t, T^{(a)}_t, T^{(b)}_t, L^{(a)}_t, L^{(b)}_t, C^{(a)}_t, C^{(b)}_t)$,
where $P^{(a)}$ (resp. $P^{(b)}$) counts the number of upward (resp. downward) price moves,
$T^{(a)}$ (resp. $T^{(b)}$) counts the number of market orders at the ask\footnote{i.e. buy orders that are executed and removed from the list } (resp. at the bid) that do not move the price,
$L^{(a)}$ (resp. $L^{(b)}$) counts the number of limit orders at the ask\footnote{i.e. buy orders added to the list }  (resp. at the bid) that do not move the price,
and $C^{(a)}$ (resp. $C^{(b)}$) counts the number of cancel orders at the ask\footnote{i.e. the number of times a limit order at the ask is canceled: in our dataset, almost 95\% of limit orders are canceled before execution.}  (resp. at the bid) that do not
move the price.
The financial data has been provided by QuantHouse EUROPE/ASIA, and consists of DAX future contracts between 01/01/2014 and 03/01/2014.

\paragraph{Baselines. }

We compare NPHC to state-of-the art baselines:
the ODE-based algorithm (ODE) by~\cite{zhou2013learning}, the Granger Causality-based algorithm (GC) by~\cite{xu2016learning},
the ADM4 algorithm (ADM4) by~\cite{zhou_zha_le_2013}, and the Wiener-Hopf-based algorithm (WH) by \cite{bacry14}.

\paragraph{Metrics.} 

We evaluate the performance of the proposed methods using the
computing time, the Relative Error
\begin{equation*}
	\text{RelErr}(\boldsymbol{A},\boldsymbol{B}) = \frac{1}{d^2} \sum_{i,j} \frac{| a^{ij} - b^{ij} |}{|a^{ij}|} \mathbbm{1}_{ \{ a^{ij} \ne 0 \}} + |b^{ij}| \mathbbm{1}_{ \{ a^{ij}=0 \}}
\end{equation*}
and the Mean Kendall Rank Correlation
\begin{equation*}
	\text{MRankCorr}(\boldsymbol{A},\boldsymbol{B}) = \frac{1}{d} \sum_{i=1}^d \text{RankCorr}([a^{i \bullet}], [b^{i \bullet}]),
\end{equation*}
where $\text{RankCorr}(x, y) = \frac{2}{ d (d-1) } ( N_{\text{concordant}}(x, y) - N_{\text{discordant}}(x, y) )$ with $N_{\text{concordant}}(x, y)$ the number of pairs $(i,j)$ satisfying $x_i > x_j$ and $y_i > y_j$ or $x_i < x_j$ and $y_i < y_j$ and $N_{\text{discordant}}(x, y)$ the number of pairs $(i,j)$ for which the same condition is not satisfied.

Note that $\text{RankCorr}$ score is a value between $-1$ and $1$, representing rank matching, but can take smaller values (in absolute value) if the entries of the vectors are not distinct.

\begin{figure}[ht]
\label{res:dax}
\begin{center}
\subfigure{\includegraphics[width=8.5cm]{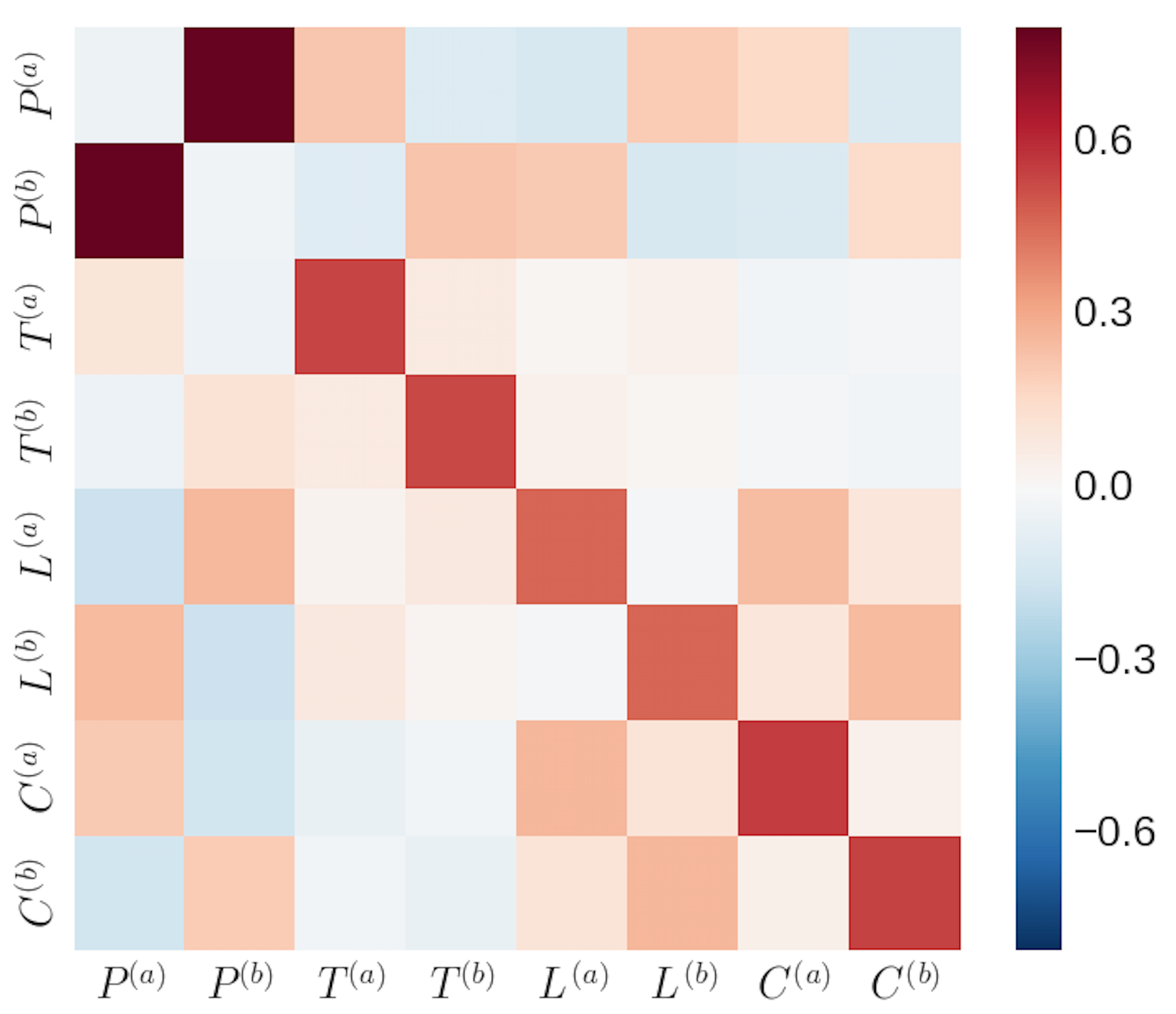}}
\caption{Estimated $\widehat{\bG}$ via NPHC on DAX order book data.}
\end{center}
\end{figure}

\begin{table}[h]
\small
  \caption{Metrics on Rect10: comparable rank correlation, strong improvement for relative error and computing time.}
  \label{sample-table}
  \centering
  \begin{tabular}{llllll}
    \toprule
    Method  & ODE & GC & ADM4 & WH & NPHC \\
    \midrule
    RelErr &0.007 &0.15 &0.10 &0.005 &\textbf{0.001}   \\
    MRankCorr  &0.33 &0.02 &0.21 &\textbf{0.34} &\textbf{0.34}   \\
    Time (s)  &846  &768  &709 &933  &\textbf{20}   \\
    \bottomrule
  \end{tabular}
\end{table}

\begin{table}[h]
\small
  \caption{Metrics on PLaw10: comparable rank correlation, strong improvement for relative error and computing time.}
  \centering
  \begin{tabular}{llllll}
    \toprule
    Method  & ODE & GC & ADM4 & WH & NPHC \\
    \midrule
    RelErr &0.011 &0.09 &0.053 &0.009 &\textbf{0.0048}   \\
    MRankCorr  &0.31 &0.26 &0.24 &\textbf{0.34} &0.33   \\
    Time (s)  &870  &781  &717 &946  &\textbf{18}   \\
    \bottomrule
  \end{tabular}
\end{table}

\begin{table}[h]
\small
  \caption{Metrics on Exp100: comparable rank correlation, strong improvement for relative error and computing time.}
  \centering
  \begin{tabular}{lllll}
    \toprule
    Method  & ODE & GC & ADM4 & NPHC \\
    \midrule
    RelErr &0.092 &0.112 &0.079 &\textbf{0.008}   \\
    MRankCorr  &0.032 &0.009 &\textbf{0.049} &0.041   \\
    Time (s)  &3215  &2950  &2411  &\textbf{47}  \\
    \bottomrule
  \end{tabular}
\end{table}

\begin{table}[h]
\small
  \caption{Metrics on MemeTracker: strong improvement in relative error, rank correlation and computing time.}
  \centering
  \begin{tabular}{lllll}
    \toprule
    Method  & ODE & GC & ADM4 & NPHC \\
    \midrule
    RelErr &0.162 &0.19 &0.092 &\textbf{0.071}   \\
    MRankCorr  &0.07 &0.053 &0.081 &\textbf{0.095}  \\
    Time (s)  &2944  &2780  &2217  &\textbf{38}  \\
    \bottomrule
  \end{tabular}
\end{table}

\paragraph{Discussion.}

We perform the ADM4 estimation, with exponential kernel, by giving the exact value $\beta=\beta_0$ of one block.
Let us stress that this helps a lot this baseline, in comparison to NPHC where nothing is specified on the shape of the kernel functions.
We used $M=10$ basis functions for both ODE and GC algorithms, and $L=50$ quadrature points for WH.
We did not run WH on the 100-dimensional datasets, for computing time reasons, because its complexity scales with $d^4$.
We ran multi-processed versions of the baseline methods on 56 cores, to decrease the computing time.

Our method consistently performs better than all baselines, on the three synthetic datasets, on MemeTracker and on the financial dataset, both in terms of Kendall rank correlation and estimation error.
Moreover, we observe that our algorithm is roughly 50 times faster than all the considered baselines.

On Rect10, PLaw10 and Exp100 our method gives very impressive results, despite the fact that it does not uses any prior shape on the kernel functions, while for instance the ADM4 baseline do.
On Figure \ref{res:nphc_exp100}, we observe that the matrix $\widehat{\bG}$ estimated with ADM4 recovers well the block for which $\beta=\beta_0$, i.e. the value we gave to the method,
but does not perform well on the two other blocks, while the matrix $\widehat{\bG}$ estimated with NPHC approximately reaches the true value for each of the three blocks.
On these simulated datasets, NPHC obtains a comparable or slightly better Kendall rank correlation, but improves a lot the relative error.

On MemeTracker, the baseline methods obtain a high relative error between 9\% and 19\% while our method achieves a relative error of 7\% which is a strong improvement.
Moreover, NPHC reaches a much better Kendall rank correlation, which proves that it leads to a much better recovery of the relative order of estimated influences than all the baselines.
Indeed, it has been shown in~\cite{zhou2013learning} that kernels of MemeTracker data are not exponential, nor power law. This partly explains why our approach behaves better.

On the financial data, the estimated kernel norm matrix obtained via NPHC, see Figure \ref{res:dax}, gave some interpretable results (see also~\cite{2014jaisson}):
\begin{enumerate}
  \item Any $2\times 2$ sub-matrix with same kind of inputs (i.e. Prices changes, Trades, Limits or Cancels) is symmetric. This shows empirically that ask and bid have symmetric roles.
  \item The prices are mostly cross-excited, which means that a price increase is very likely to be followed by a price decrease, and conversely. This is consistent with the wavy prices we observe on financial markets.
  \item The market, limit and cancel orders are strongly self-excited. This can be explained by the persistence of order flows, and by the splitting of meta-orders into sequences of smaller orders. Moreover, we observe that orders impact the price without changing it. For example, the increase of cancel orders at the bid causes downward price moves.
\end{enumerate}

\section{Technical details}
\label{sec:proofs}

We show in this section how to obtain the equations stated above, the estimators of the integrated cumulants and the scaling coefficient $\kappa$ that appears in the objective function. We then prove the theorem of the paper.

\subsection{Proof of Equation~\eqref{eq:relation1}}

We denote $\boldsymbol{\nu}(z)$ the matrix
\begin{equation*}
	\nu^{ij}(z) =\mathcal{L}_z\Big(t \rightarrow
\frac{\mathbb{E}(dN_{u}^{i} dN_{u+t}^{j})}{du dt} - \Lambda^i \Lambda^j \Big),
\end{equation*}
where $\mathcal{L}_z(f)$ is the Laplace transform of $f$, and $\psi_t = \sum_{n \ge 1} \phi^{(\star n)}_t$, where $\phi^{(\star n)}_t$ refers to the $n^{th}$ auto-convolution of $\phi_t$.
Then we use the characterization of second-order statistics, first formulated in \cite{hawkes1971point} and fully generalized in \cite{bacry14},
\begin{equation*}
\boldsymbol{\nu}(z) = (\bI + \mathcal{L}_{-z}(\boldsymbol{\Psi})) \bL (\bI + \mathcal{L}_z(\boldsymbol{\Psi}))^\top,
\end{equation*}
where $\boldsymbol L^{ij} = \Lambda^i \delta^{ij}$ with $\delta^{ij}$ the Kronecker symbol.
Since $\bI + \mathcal{L}_z(\boldsymbol{\Psi}) = (\bI - \mathcal{L}_z(\boldsymbol{\Phi}))^{-1}$, taking $z=0$ in the previous equation gives
\begin{align*}
\boldsymbol{\nu}(0) &= (\bI - \bG)^{-1} \bL (\bI - \bG^\top)^{-1}, \\
\bC &= \bR \bL \bR^\top,
\end{align*}
which gives us the result since the entry $(i,j)$ of the last equation gives $C^{ij} = \sum_m \Lambda^m R_{im} R_{jm}$.

\subsection{Proof of Equation~\eqref{eq:relation2}}

We start from~\cite{hk_cumul}, cf. Eqs.~(48) to~(51), and group some terms:
\begin{align*}
K^{ijk} &= \sum_m \Lambda^m R_{im} R_{jm} R_{km} \\
&+ \sum_m R_{im} R_{jm} \sum_n \Lambda^n R_{kn} \mathcal{L}_0(\psi^{mn}) \\
&+ \sum_m R_{im} R_{km} \sum_n \Lambda^n R_{jn} \mathcal{L}_0(\psi^{mn}) \\
&+ \sum_m R_{jm} R_{km} \sum_n \Lambda^n R_{in} \mathcal{L}_0(\psi^{mn}).
\end{align*}
Using the relations $\mathcal{L}_0(\psi^{mn}) = R_{mn} - \delta^{mn}$ and $C^{ij} = \sum_m \Lambda^m R_{im} R_{jm}$, proves Equation~\eqref{eq:relation2}.

\subsection{Integrated cumulants estimators}
\label{appen:estimators}

For $H>0$ let us denote $\Delta_H N^i_t = N^i_{t+H}-N^i_{t-H}$.
Let us first remark that, if one restricts the integration domain to $(-H,H)$
in Eqs. \eqref{eq:cumul2density} and \eqref{eq:cumul3density}, one gets by permuting integrals and expectations:
\begin{align*}
\Lambda^i dt &=  \mathbb{E}(dN_t^i) \\
C^{ij} dt &= \mathbb{E} \left(dN_{t}^i (\Delta_H N^j_{t}- 2 H \Lambda^j) \right)\\
K^{ijk} dt &= \mathbb{E} \left(dN_{t}^i (\Delta_H N^j_{t}- 2 H \Lambda^j)(\Delta_H N^k_t -2 H \Lambda^k) \right) \\
& \quad -
dt \Lambda^i \mathbb{E}\left( (\Delta_H N^j_t-2H \Lambda^j)(\Delta_H N^k_t-2H \Lambda^k) \right).
\end{align*}
The estimators \eqref{eq:estimator1} and \eqref{eq:estimator2}
are then naturally obtained by replacing the expectations by their empirical counterparts, notably
\begin{equation*}
	\frac{\mathbb{E}(dN^i_t f(t))}{dt} \rightarrow \frac{1}{T} \sum_{\tau \in Z^i} f(\tau).
\end{equation*}
For the estimator~\eqref{eq:estimator3}, we shall also notice that
\begin{align*}
  \mathbb{E}( &(\Delta_H N^j_t-2H \Lambda^j)(\Delta_H N^k_t-2H \Lambda^k) ) \\
  &= \int \int \mathbbm{1}_{[-H,H]}(t) \mathbbm{1}_{[-H,H]}(t') C^{jk}_{t-t'}dtdt' \\
  &= \int (2H - |t|)^{+} C^{jk}_t dt.
\end{align*}
We estimate the last integral with the remark above.

\subsection{Choice of the scaling coefficient $\kappa$}
\label{sec:kappa}

Following the theory of GMM, we denote $m(X,\theta)$ a function of the data, where $X$ is distributed with respect to a distribution $\mathbb{P}_{\theta_0}$,
which satisfies the \emph{moment conditions} $g(\theta) = \mathbb{E}[m(X,\theta)] = 0$ if and only if $\theta = \theta_0$, the parameter $\theta_0$ being the \emph{ground truth}.
For $x_1, \ldots, x_N$ observed copies of $X$, we denote $\widehat{g}_i (\theta) = m(x_i, \theta)$,
the usual choice of weighting matrix is $\widehat{W}_N (\theta) = \frac{1}{N} \sum_{i=1}^N \widehat{g}_i (\theta) \widehat{g}_i (\theta)^\top$,
and the objective to minimize is then
\begin{align}
\left( \frac{1}{N} \sum_{i=1}^N \widehat{g}_i (\theta) \right) \left( \widehat{W}_N (\theta_1)\right)^{-1} \left( \frac{1}{N} \sum_{i=1}^N \widehat{g}_i (\theta) \right),
\end{align}
where $\theta_1$ is a constant vector. Instead of computing the inverse weighting matrix, we rather use its projection on $\{ \alpha \bI: \alpha \in \mathbb{R} \}$.
It can be shown that the projection choses $\alpha$ as the mean eigenvalue of $\widehat{W}_N (\theta_1)$. We can easily compute the sum of its eigenvalues:
\begin{align*}
\mbox{Tr}(\widehat{W}_N (\theta_1)) &= \frac{1}{N} \sum_{i=1}^N \mbox{Tr}(\widehat{g}_i (\theta_1) \widehat{g}_i (\theta_1)^\top) = \frac{1}{N} \sum_{i=1}^N \mbox{Tr}(\widehat{g}_i (\theta_1)^\top \widehat{g}_i (\theta_1)) = \frac{1}{N} \sum_{i=1}^N || \widehat{g}_i (\theta_1)||^2_2.
\end{align*}
In our case, $\widehat{g} (\bR) = \left[ \mbox{\textbf{vec}} [ \widehat{\bKc} - \bKc(\bR) ], \mbox{\textbf{vec}} [ \widehat{\bC} - \bC (\bR) ] \right]^\top \in \mathbb{R}^{2 d^2}$.
Considering a block-wise weighting matrix, one block for $\widehat{\bKc} - \bKc(\bR)$ and the other for $\widehat{\bC} - \bC (\bR)$,
the sum of the eigenvalues of the first block becomes $\|\widehat{\bKc} - \bKc(\bR)\|^2_2$, and $\| \widehat{\bC} - \bC (\bR) \|^2_2$ for the second.
We compute the previous terms with $\bR_1 = 0$. All together, the objective function to minimize is
\begin{align}
 \frac{1}{\| \widehat{\bKc} \|_2^2} \|\bKc(\bR) - \widehat{\bKc}\|_2^2 + \frac{1}{\| \widehat{\bC} \|_2^2} \| \bC (\bR) - \widehat{\bC} \|_2^2.
\end{align}
Dividing this function by $\left( 1 / \| \widehat{\bKc} \|_2^2 + 1 / \| \widehat{\bC} \|_2^2 \right)^{-1}$, and setting $\kappa = \|\widehat{\bKc}\|^2_2 / (\|\widehat{\bKc}\|^2_2 +
 \|\widehat{\bC}\|^2_2)$,
we obtaind the loss function given in Equation (\ref{eq:nphc_loss}).

\subsection{Proof of the Theorem}
\label{sec:proof_theorem}

The main difference with the usual Generalized Method of Moments, see \cite{hansen1982large}, relies in the relaxation of the moment conditions,
since we have $\mathbb{E}[\widehat{g}_T(\theta_0)] = m_T \ne 0$.
We adapt the proof of consistency given in \cite{newey1994large}. \\

We can relate the integral of the Hawkes process's kernels to the integrals of the cumulant densities, from \cite{hk_cumul}.
Our cumulant matching method would fall into the usual GMM framework if we could estimate - without bias - the integral of the covariance on $\mathbb{R}$, and the integral of the skewness on $\mathbb{R}^2$.
Unfortunately, we can't do that easily. We can however estimate without bias $\int f^T_t C^{ij}_t dt$ and $\int f^T_t K^{ijk}_t dt$
with $f^T$ a compact supported function on $[-H_T,H_T]$ that weakly converges to $1$, with $H_T \longrightarrow \infty$. 
In most cases we will take $f^T_t = \mathbbm{1}_{[-H_T,H_T]}(t)$.
Denoting $\widehat{C}^{ij,(T)}$ the estimator of $\int f^T_t C^{ij}_t dt$, the term $| \mathbb{E}[\widehat{C}^{ij,(T)}] - C^{ij} | = | \int f^T_t C^{ij}_t dt - C^{ij} |$
can be considered a proxy to the \emph{distance to the classical GMM}.
This distance has to go to zero to make the rest of GMM's proof work: the estimator $\widehat{C}^{ij,(T)}$ is then asymptotically unbiased towards $C^{ij}$ when $T$ goes to infinity.

\subsubsection{Notations}
We observe the multivariate point process $(\bN_t)$ on $\mathbb{R}^{+}$, with $Z^i$ the events of the $i^{th}$ component.
We will often write covariance / skewness instead of integrated covariance / skewness.
In the rest of the document, we use the following notations. \\
\\
\noindent \textbf{Hawkes kernels' integrals} $\quad \bG^{\text{true}} = \int \boldsymbol{\Phi}_t dt = (\int \phi^{ij}_t dt)_{ij} = \bI - (\bR^{\text{true}})^{-1}$ \\
\\
\noindent \textbf{Theoretical mean matrix} $\quad \bL = \mbox{diag}(\Lambda^1, \ldots, \Lambda^d)$ \\
\\
\noindent \textbf{Theoretical covariance} $\quad \bC = \bR^{\text{true}} \bL (\bR^{\text{true}})^\top$ \\
\\
\noindent \textbf{Theoretical skewness} $\quad \bKc = (K^{iij})_{ij} = (\bR^{\text{true}})^{\odot^2} \bC^\top + 2 [\bR^{\text{true}} \odot (\bC - \bR^{\text{true}} \bL)] (\bR^{\text{true}})^\top$ \\
\\
\noindent \textbf{Filtering function} $\quad f^T \ge 0 \quad \quad \mbox{supp}(f^T) \subset [-H_T,H_T] \quad \quad F^T = \int f^T_s ds \quad \quad \widetilde{f}^T_t = f^T_{-t}$ \\
\\
\noindent \textbf{Events sets} $\quad Z^{i,T,1} = Z^i \cap [H_T,T+H_T] \quad \quad \quad Z^{j,T,2} = Z^j\cap[0,T+2H_T]$ \\
\\
\noindent \textbf{Estimators of the mean} $\quad \widehat{\Lambda}^i = \frac{N^i_{T+H_T} - N^i_{H_T}}{T} \quad \quad \quad \widetilde{\Lambda}^j = \frac{N^j_{T+2 H_T}}{T+2 H_T}$ \\
\\
\noindent \textbf{Estimator of the covariance} $\quad \widehat{C}^{ij,(T)} = \frac{1}{T} \sum_{\tau \in Z^{i,T,1}} \left( \sum_{\tau' \in Z^{j,T,2}} f_{\tau'-\tau} - \widetilde{\Lambda}^j F^T \right)$ \\
\\
\noindent \textbf{Estimator of the skewness\footnote{When $f^T_t = \mathbbm{1}_{[-H_T,H_T]}(t)$, we remind that $(f^T \star \widetilde{f}^T)_t = (2 H_T - |t|)^{+}$. This leads to the estimator we showed in the article.}}
\begin{align*}
\quad \widehat{K}^{ijk,(T)} &= \frac{1}{T} \sum_{\tau \in Z^{i,T,1}} \left( \sum_{\tau' \in Z^{j,T,2}} f_{\tau'-\tau} - \widetilde{\Lambda}^j F^T \right) \left( \sum_{\tau'' \in Z^{k,T,2}} f_{\tau' - \tau} - \widetilde{\Lambda}^k F^T \right) \\
&- \frac{\widehat{\Lambda}^i}{T + 2 H_T} \sum_{\tau' \in Z^{j,T,2}} \left( \sum_{\tau'' \in Z^{k,T,2}} (f^T \star \widetilde{f}^T)_{\tau'-\tau''} - \widetilde{\Lambda}^k (F^T)^2 \right)
\end{align*}
\subsubsection*{GMM related notations}
\begin{align*}
  \theta &= \bR \quad \mbox{ and } \quad \theta_0 = \bR^{\text{true}}\\
  g_0 (\theta) &= \textsf{\textbf{vec}} \begin{bmatrix}
     \bC - \bR \bL \bR^\top \\
     \bKc - \bR^{\odot^2} \bC^\top - 2 [\bR \odot (\bC - \bR \bL)] \bR^\top
  \end{bmatrix} \in \mathbb{R}^{2d^2} \\
\widehat{g}_T (\theta) &= \textsf{\textbf{vec}} \begin{bmatrix}
\widehat{\bC}^{(T)} - \bR \widehat{\bL} \bR^\top \\
\widehat{\bKc}^{(T)} - \bR^{\odot^2} (\widehat{\bC}^{(T)})^\top - 2 [\bR \odot (\widehat{\bC}^{(T)} - \bR \widehat{\bL})] \bR^\top
\end{bmatrix} \in \mathbb{R}^{2d^2} \\
Q_0 (\theta) &= g_0 (\theta)^\top W g_0 (\theta) \\
\widehat{Q}_T (\theta) &= \widehat{g}_T (\theta)^\top \widehat{W}_T \widehat{g}_T (\theta)
\end{align*}

\subsubsection{Consistency}
First, let's remind a useful theorem for consistency in GMM from \cite{newey1994large}.
\begin{theorem}
If there is a function $Q_0 (\theta)$ such that
$(i)$~$Q_0 (\theta)$ is uniquely maximized at $\theta_0$;
$(ii)$~$\Theta$ is compact;
$(iii)$~$Q_0(\theta)$ is continuous;
$(iv)$~$\widehat{Q}_T (\theta)$ converges uniformly in probability to $Q_0 (\theta)$,
then $\widehat{\theta}_T = \arg \max \widehat{Q}_T (\theta) \overset{\mathbb{P}}{\longrightarrow} \theta_0$.
\end{theorem}
\noindent We can now prove the consistency of our estimator.
\begin{theorem}
Suppose that $(\boldsymbol{N_t})$ is observed on $\mathbb{R}^{+}$, $\widehat{W}_T \overset{\mathbb{P}}{\longrightarrow} W$, and
\begin{enumerate}
\item $W$ is positive semi-definite and $W g_0 (\theta) = 0$ if and only if $\theta=\theta_0$,
\item $\theta \in \Theta$, which is compact,
\item the spectral radius of the kernel norm matrix satisfies $||\boldsymbol{\Phi}||_{*} < 1$,
\item $\forall i, j, k \in [d], \int f_u^T C^{ij}_u du \rightarrow \int C^{ij}_u du$ and $\int f_u^T f_v^T K^{ijk}_{u,v} du dv \rightarrow \int K^{ijk}_{u,v} du dv$,
\item $(F^T)^2 / T \overset{\mathbb{P}}{\longrightarrow} 0$ and $||f||_\infty = O(1)$.
\end{enumerate}
Then
\begin{align*}
  \widehat{\theta}_T \overset{\mathbb{P}}{\longrightarrow} \theta_0.
\end{align*}
\end{theorem}

\begin{remark}
  In practice, we use a constant sequence of weighting matrices: $\widehat{W}_T = \bI$.
\end{remark}

\begin{proof}
Proceed by verifying the hypotheses of Theorem 2.1 from \cite{newey1994large}.
Condition 2.1$(i)$ follows by $(i)$ and by $Q_0(\theta) = [W^{1/2} g_0(\theta)]^\top [W^{1/2} g_0(\theta)] > 0 = Q_0(\theta_0)$.
Indeed, there exists a neighborhood $N$ of $\theta_0$ such that $\theta \in N \backslash \{\theta_0\} $ and $g_0(\theta) \ne 0$ since $g_0(\theta)$ is a polynom.
Condition 2.1$(ii)$ follows by $(ii)$.
Condition 2.1$(iii)$ is satisfied since $Q_0 (\theta)$ is a polynom.
Condition 2.1$(iv)$ is harder to prove.
First, since $\widehat{g}_T (\theta)$ is a polynom of $\theta$, we prove easily that $\mathbb{E}[\sup_{\theta \in \Theta} | \widehat{g}_T (\theta) |] < \infty$.
Then, by $\Theta$ compact, $g_0 (\theta)$ is bounded on $\Theta$, and by the triangle and Cauchy-Schwarz inequalities,
\begin{align*}
\big| \widehat{Q}_T & (\theta) - Q_0 (\theta) \big| \\
&\le \big| ( \widehat{g}_T (\theta) - g_0 (\theta) )^\top \widehat{W}_T ( \widehat{g}_T (\theta) - g_0 (\theta) ) \big| \\
&+ \big| g_0 (\theta)^\top (\widehat{W}_T + \widehat{W}_T^\top) ( \widehat{g}_T (\theta) - g_0 (\theta) ) \big|
+ \big| g_0 (\theta)^\top (\widehat{W}_T - W) g_0 (\theta) \big| \\
&\le \norm{\widehat{g}_T (\theta) - g_0 (\theta)}^2 \norm{\widehat{W}_T}
+ 2 \norm{ g_0 (\theta) } \norm{ \widehat{g}_T (\theta) - g_0 (\theta) } \norm{ \widehat{W}_T }
+ \norm{ g_0 (\theta) }^2 \norm{ \widehat{W}_T - W }.
\end{align*}
To prove $\sup_{\theta \in \Theta} \big| \widehat{Q}_T (\theta) - Q_0 (\theta) \big| \overset{\mathbb{P}}{\longrightarrow} 0$,
we should now prove that $\sup_{\theta \in \Theta} \norm{\widehat{g}_T (\theta) - g_0 (\theta)} \overset{\mathbb{P}}{\longrightarrow} 0$.
By $\Theta$ compact, it is sufficient to prove that
$\norm{\widehat{\bL} - \bL}~\overset{\mathbb{P}}{\longrightarrow}~0$,
$\norm{\widehat{\bC}^{(T)} - \bC}~\overset{\mathbb{P}}{\longrightarrow}~0$,
and $\norm{\widehat{\bKc}^{(T)} - \bKc}~\overset{\mathbb{P}}{\longrightarrow}~0$.

\subsubsection*{Proof that $\norm{\widehat{\bL} - \bL}~\overset{\mathbb{P}}{\longrightarrow}~0$}
The estimator of $\bL$ is unbiased so let's focus on the variance of $\widehat{\bL}$.
\begin{align*}
\mathbb{E}[(\widehat{\Lambda}^i - \Lambda^i)^2] &= \mathbb{E}\left[ \left( \frac{1}{T} \int_{H_T}^{T+H_T} (dN^i_t - \Lambda^i dt) \right)^2 \right] \\
&= \frac{1}{T^2} \int_{H_T}^{T+H_T} \int_{H_T}^{T+H_T} \mathbb{E} [ (dN^i_t - \Lambda^i dt) (dN^i_{t'} - \Lambda^i dt') ] \\
&= \frac{1}{T^2} \int_{H_T}^{T+H_T} \int_{H_T}^{T+H_T} C^{ii}_{t'-t} dt dt' \\
&\le \frac{1}{T^2} \int_{H_T}^{T+H_T} C^{ii} dt = \frac{C^{ii}}{T} \longrightarrow 0
\end{align*}
By Markov inequality, we have just proved that $\norm{\widehat{\bL} - \bL}~\overset{\mathbb{P}}{\longrightarrow}~0$.

\subsubsection*{Proof that $\norm{\widehat{\bC}^{(T)} - \bC}~\overset{\mathbb{P}}{\longrightarrow}~0$}
First, let's remind that $\mathbb{E}(\widehat{\bC}^{(T)}) \ne \bC$.
Indeed,
\begin{align*}
\mathbb{E}\left(\widehat{C}^{ij,(T)}\right) &= \mathbb{E}\left( \frac{1}{T} \int_{H_T}^{T+H_T} dN^i_t \int_{0}^{T+2H_T} dN^j_{t'} f_{t'-t} - \widehat{\Lambda}^i \widetilde{\Lambda}^j F^T \right) \\
&= \mathbb{E}\left( \frac{1}{T} \int_{H_T}^{T+H_T} dN^i_t \int_{-t}^{T+2H_T-t} dN^j_{t+s} f_{s} - \Lambda^i \Lambda^j F^T \right) + \epsilon^{ij,T,H_T} F^T \\
&= \frac{1}{T} \int_{H_T}^{T+H_T} \int_{-H_T}^{H_T} f_s \mathbb{E}\left( dN^i_t dN^j_{t+s} - \Lambda^i \Lambda^j  ds \right) + \epsilon^{ij,T,H_T} F^T \\
&= \int f_s C^{ij}_s ds + \epsilon^{ij,T,H_T} F^T
\end{align*}
Now,
\begin{align*}
\epsilon^{ij,T,H_T} &= \mathbb{E} \left( \Lambda^i \Lambda^j - \widehat{\Lambda}^i \widetilde{\Lambda}^j \right) \\
&= - \frac{1}{T^2} \int_{H_T}^{T+H_T} \int_{0}^{T+2H_T} \mathbb{E} \left( dN^i_t dN^j_{t'} - \Lambda^i \Lambda^j dt dt' \right)\\
&= - \frac{1}{T^2} \int_{H_T}^{T+H_T} \int_{0}^{T+2H_T} C^{ij}_{t-t'} dt dt' \\
&= - \frac{1}{T} \int \left( 1 + \left(\frac{H_T-|t|}{T}\right)^{-} \right)^{+}  C^{ij}_t dt
\end{align*}
Since $f$ satisfies $F^T = o(T)$, we have $\mathbb{E}(\widehat{\bC}^{(T)}) \longrightarrow \bC$.
It remains now to prove that $\norm{\widehat{\bC}^{(T)} - \mathbb{E}(\widehat{\bC}^{(T)})}~\overset{\mathbb{P}}{\longrightarrow}~0$. \\
Let's now focus on the variance of $\widehat{C}^{ij,(T)}: \quad \mathbb{V}(\widehat{C}^{ij, (T)}) = \mathbb{E}\left( (\widehat{C}^{ij, (T)})^2 \right) - \mathbb{E}(\widehat{C}^{ij, (T)})^2$. \\
Now,
\begin{align*}
& \mathbb{E} \left( (\widehat{C}^{ij, (T)})^2 \right) \\
&= \mathbb{E} \left( \frac{1}{T^2} \sum_{(\tau, \eta, \tau', \eta') \in (Z^{i,T,1})^2 \times (Z^{j,T,2})^2} ( f_{\tau'-\tau} - F^T / (T+2H_T) ) ( f_{\eta'-\eta} - F^T / (T+2H_T) ) \right) \\
&= \mathbb{E} \left( \frac{1}{T^2} \int_{t,s \in [H_T,T+H_T]} \int_{t',s'} dN^i_t dN^j_{t'} dN^i_s dN^j_{s'} ( f_{t'-t} - F^T / (T+2H_T) ) ( f_{s'-s} - F^T / (T+2H_T) ) \right) \\
&= \frac{1}{T^2} \int_{t,s \in [H_T,T+H_T]} \int_{t',s' \in [0,T+2H_T]} \mathbb{E} \left( dN^i_t dN^j_{t'} dN^i_s  dN^j_{s'} \right) \\
& \quad \quad \quad \quad \quad \quad \quad \quad \quad \quad \quad \cdot ( f_{t'-t} - F^T / (T+2H_T) ) ( f_{s'-s} - F^T / (T+2H_T) )
\end{align*}
And,
\begin{align*}
& \mathbb{E}(\widehat{C}^{ij, (T)})^2 \\
&= \frac{1}{T^2} \int_{t,s \in [H_T,T+H_T]} \int_{t',s' \in [0,T+2H_T]} \mathbb{E} \left( dN^i_t dN^j_{t'} \right) \mathbb{E} \left( dN^i_s dN^j_{s'} \right) \\
& \quad \quad \quad \quad \quad \quad \quad \quad \quad \quad \quad \cdot ( f_{t'-t} - F^T / (T+2H_T) ) ( f_{s'-s} - F^T / (T+2H_T) )
\end{align*}
Then, the variance involves the integration towards the difference of moments $\mu^{r,s,t,u} - \mu^{r,s} \mu^{t,u}$.
Let's write it as a sum of cumulants, since cumulants density are integrable.
\begin{align*}
\mu^{r,s,t,u} - \mu^{r,s} \mu^{t,u} &= \kappa^{r,s,t,u} + \kappa^{r,s,t} \kappa^u [4] + \kappa^{r,s} \kappa^{t,u} [3] + \kappa^{r,s} \kappa^{t} \kappa^{u} [6] + \kappa^r \kappa^s \kappa^t \kappa^u - (\kappa^{r,s} + \kappa^r \kappa^s ) (\kappa^{t,u} + \kappa^t \kappa^u ) \\
&= \kappa^{r,s,t,u} \\
&+ \kappa^{r,s,t} \kappa^u + \kappa^{u,r,s} \kappa^t + \kappa^{t,u,r} \kappa^s + \kappa^{s,t,u} \kappa^r \\
&+ \kappa^{r,t} \kappa^{s,u} + \kappa^{r,u} \kappa^{s,t} \\
&+ \kappa^{r,t} \kappa^{s} \kappa^{u} + \kappa^{r,u} \kappa^{s} \kappa^{t} + \kappa^{s,t} \kappa^{r} \kappa^{u} + \kappa^{s,t} \kappa^{r} \kappa^{u}
\end{align*}
In the rest of the proof, we denote $a_t = \mathbbm{1}_{ t \in [H_T,T+H_T] }$, $b_t = \mathbbm{1}_{ t \in [0,T+2H_T] }$, $c_t = \mathbbm{1}_{ t \in [-H_T,H_T] }$, $g_t = f_t - \frac{1}{T+2H_T} F^T$ \\
Before starting the integration of each term, let's remark that:
\begin{enumerate}
\item $\boldsymbol{\Psi}_t = \sum_{n \ge 1} \boldsymbol{\Phi}^{(\star n)}_t \ge 0$ since $\boldsymbol{\Phi}_t \ge 0$.
\item The regular parts of $C^{ij}_u$, $K^{ijk}_{u,v}$ (skewness density) and $M^{ijkl}_{u,v,w}$ (fourth cumulant density) are positive as polynoms of integrals of $\psi^{ab}_\cdot$ with positive coefficients. The integrals of the singular parts are positive as well.
\item
\begin{enumerate}
\item $\int a_t b_{t'} f_{t'-t} dt dt' = T F^T$
\item $\int a_t b_{t'} g_{t'-t} dt dt' = 0$
\item $\int a_t b_{t'} |g_{t'-t}| dt dt' \le 2 T F^T$
\end{enumerate}
\item $\forall t \in \mathbb{R}, a_t (b \star \widetilde{g})_t = 0$, where $\widetilde{g}_s = g_{-s}$.
\end{enumerate}
\noindent \textbf{Fourth cumulant} $\quad$
We want here to compute $\int \kappa^{i,j,i,j}_{t,t',s,s'} a_{t} b_{t'} a_{s} b_{s'} g_{t'-t} g_{s'-s} dt dt' ds ds'$. \\
We remark that $|g_{t'-t} g_{s'-s}| \le (||f||_\infty (1 + 2H_T/T))^2 \le 4 ||f||_\infty^2$.
\begin{align*}
\Big| \frac{1}{T^2} \int \kappa^{i,j,i,j}_{t,t',s,s'} a_{t} b_{t'} a_{s} b_{s'} g_{t'-t} g_{s'-s} dt dt' ds ds' \Big| &\le \left( \frac{2 ||f||_\infty}{T} \right)^2 \int dt a_t \int dt' b_{t'} \int ds a_s \int ds' b_{s'} M^{ijij}_{t'-t,s-t,s'-t}\\
&\le \left( \frac{2 ||f||_\infty}{T} \right)^2 \int dt a_t \int dt' b_{t'} \int ds a_s \int dw M^{ijij}_{t'-t,s-t,w} \\
&\le \left( \frac{2 ||f||_\infty}{T} \right)^2 \int dt a_t \int M^{ijij}_{u,v,w} du dv dw \\
&\le \frac{4 ||f||_\infty^2}{T} M^{ijij} \underset{T \rightarrow \infty}{\longrightarrow} 0
\end{align*}

\noindent \textbf{Third $\times$ First} $\quad$
We have four terms, but only two different forms since the roles of $(s,s')$ and $(t,t')$ are symmetric. \\
\noindent First form
\begin{align*}
\int \kappa^{i,j,i}_{t,t',s} \Lambda^j G_{\boldsymbol{t}} d\boldsymbol{t} &= \frac{\Lambda^j}{T^2} \int \kappa^{i,j,i}_{t,t',s} a_{t} b_{t'} a_{s} b_{s'} g_{t'-t} g_{s'-s} dt dt' ds ds' \\
&= \frac{\Lambda^j}{T^2} \int \kappa^{i,j,i}_{t,t',s} a_{t} b_{t'} a_{s} (b \star \widetilde{g})_{s} g_{t'-t} dt dt' ds \\
&= 0 \quad \quad \mbox{ since } a_{s} (b \star \widetilde{g})_{s} = 0
\end{align*}

\noindent Second form
\begin{align*}
\Big| \int \kappa^{i,j,j}_{t,t',s'} \Lambda^i G_{\boldsymbol{t}} d\boldsymbol{t} \Big| &= \Big| \frac{\Lambda^i}{T^2} \int \kappa^{i,j,j}_{t,t',s'} a_{t} b_{t'} a_{s} b_{s'} g_{t'-t} g_{s'-s} dt dt' ds ds' \Big| \\
&= \Big| \frac{\Lambda^i}{T^2} \int \kappa^{i,j,j}_{t,t',s'} a_{t} b_{t'} g_{t'-t} b_{s'} (a \star g)_{s'} dt dt' ds' \Big| \\
&\le \frac{\Lambda^i}{T^2} 2 ||f||_\infty \int ds' b_{s'} (a \star |g|)_{s'} \int dt a_t \int dt' b_{t'} K^{ijj}_{t'-s', t-s'} \\
&\le 4 ||f||_\infty K^{ijj} \Lambda^i \frac{F^T}{T} \underset{T \rightarrow \infty}{\longrightarrow} 0
\end{align*}

\noindent \textbf{Second $\times$ Second} $\quad$ \\
\noindent First form
\begin{align*}
\Big| \int \kappa^{i,i}_{t,s} \kappa^{j,j}_{t',s'} G_{\boldsymbol{t}} d\boldsymbol{t} \Big| &\le \frac{2 ||f||_\infty}{T^2} \int C^{ii}_{t-s} C^{jj}_{t'-s'} a_t b_{t'} |g_{t'-t}| a_s b_{s'} dt dt' ds ds' \\
&\le \frac{2 ||f||_\infty}{T^2} C^{ii} C^{jj} \int a_t b_{t'} |g_{t'-t}| dt dt' \\
&\le 4 ||f||_\infty C^{ii} C^{jj} \frac{F^T}{T} \underset{T \rightarrow \infty}{\longrightarrow} 0
\end{align*}

\noindent Second form
\begin{align*}
\Big| \int \kappa^{i,j}_{t,s'} \kappa^{i,j}_{t',s} G_{\boldsymbol{t}} d\boldsymbol{t} \Big| &\le 4 ||f||_\infty (C^{ij})^2 \frac{F^T}{T} \underset{T \rightarrow \infty}{\longrightarrow} 0
\end{align*}

\noindent \textbf{Second $\times$ First $\times$ First} $\quad$ \\
\noindent First form
\begin{align*}
\int \kappa^{i,j}_{t,t'} \Lambda^i \Lambda^j G_{\boldsymbol{t}} d\boldsymbol{t} &= \frac{\Lambda^i \Lambda^j}{T^2} \int \kappa^{i,j}_{t,t'} a_{t} b_{t'} g_{t'-t} dt dt' \int a_{s} b_{s'} g_{s'-s} ds ds' = 0
\end{align*}
\noindent Second form
\begin{align*}
\int \kappa^{i,i}_{t,s} \Lambda^j \Lambda^j G_{\boldsymbol{t}} d\boldsymbol{t} &= \left( \frac{\Lambda^j}{T} \right)^2 \int \kappa^{i,i}_{t,s} a_{t} b_{t'} g_{t'-t} a_{s} (b \star \widetilde{g})_{s} dt dt' ds = 0
\end{align*}
\noindent We have just proved that $\mathbb{V}(\widehat{\bC}^{(T)}) \overset{\mathbb{P}}{\longrightarrow} 0$.
By Markov inequality, it ensures us that $\norm{\widehat{\bC}^{(T)} - \mathbb{E}(\widehat{\bC}^{(T)})}~\overset{\mathbb{P}}{\longrightarrow}~0$,
and finally that $\norm{\widehat{\bC}^{(T)} - \bC}~\overset{\mathbb{P}}{\longrightarrow}~0$.
\end{proof}

\subsubsection*{Proof that $\norm{\widehat{\bKc}^{(T)} - \bKc}~\overset{\mathbb{P}}{\longrightarrow}~0$}
The scheme of the proof is similar to the previous one.
The upper bounds of the integrals involve the same kind of terms,
plus the new term $(F^T)^2/T$ that goes to zero thanks to the assumption 5 of the theorem.

\section{Conclusion} 
\label{sec:conclusion}


In this paper, we introduce a simple nonparametric method
(the NPHC algorithm) that leads to a fast and robust estimation of the matrix $\bG$ of the kernel integrals of a Multivariate Hawkes process that encodes Granger causality between nodes.
This method relies on the matching of the integrated order 2 and order 3 empirical cumulants, which represent the simplest set of global observables
containing sufficient information to recover the matrix $\bG$.
Since this matrix fully accounts for the self- and cross- influences
of the process nodes (that can represent agents or users in applications),
our approach can naturally be used to quantify the degree of endogeneity of a system and to uncover the causality
structure of a network.

By performing numerical experiments involving very different kernel shapes, we show that
the baselines, involving either parametric or non-parametric approaches are very sensible
to model misspecification, do not lead to accurate estimation, and are numerically expensive,
while NPHC provides fast, robust and reliable results.
This is confirmed on the MemeTracker database, where we show that NPHC outperforms classical approaches based on EM algorithms or the Wiener-Hopf equations.
Finally, the NPHC algorithm provided very satisfying results on financial data, that are consistent with well-known stylized facts in finance.

\section*{Acknowledgements}
This work benefited from the support of the chair ``Changing markets'', CMAP
\'Ecole Polytechnique and \'Ecole Polytechnique fund raising - Data Science Initiative.

The authors want to thank Marcello Rambaldi for fruitful discussions on order book data's experiments.
\clearpage

\appendix

\bibliography{biblio.bib}

\begin{thebibliography}{34}
\providecommand{\natexlab}[1]{#1}
\providecommand{\url}[1]{\texttt{#1}}
\expandafter\ifx\csname urlstyle\endcsname\relax
  \providecommand{\doi}[1]{doi: #1}\else
  \providecommand{\doi}{doi: \begingroup \urlstyle{rm}\Url}\fi

\bibitem[Abadi et~al.(2016)Abadi, Agarwal, Barham, Brevdo, Chen, Citro,
  Corrado, Davis, Dean, Devin, et~al.]{abadi2016tensorflow}
M.~Abadi, A.~Agarwal, P.~Barham, E.~Brevdo, Z.~Chen, C.~Citro, G.~S. Corrado,
  A.~Davis, J.~Dean, M.~Devin, et~al.
\newblock Tensorflow: Large-scale machine learning on heterogeneous distributed
  systems.
\newblock \emph{arXiv preprint arXiv:1603.04467}, 2016.

\bibitem[A{\"\i}t-Sahalia et~al.(2010)A{\"\i}t-Sahalia, Cacho-Diaz, and
  Laeven]{ait2010modeling}
Y.~A{\"\i}t-Sahalia, J.~Cacho-Diaz, and R.~JA Laeven.
\newblock Modeling financial contagion using mutually exciting jump processes.
\newblock Technical report, National Bureau of Economic Research, 2010.

\bibitem[{Bacry} and {Muzy}(2016)]{bacry14}
E.~{Bacry} and J.-F. {Muzy}.
\newblock First- and second-order statistics characterization of hawkes
  processes and non-parametric estimation.
\newblock \emph{IEEE Transactions on Information Theory}, 62\penalty0
  (4):\penalty0 2184--2202, 2016.

\bibitem[Bacry et~al.(2015)Bacry, Mastromatteo, and Muzy]{bacry2015hawkes}
E.~Bacry, I.~Mastromatteo, and J.-F. Muzy.
\newblock Hawkes processes in finance.
\newblock \emph{Market Microstructure and Liquidity}, 1\penalty0 (01):\penalty0
  1550005, 2015.

\bibitem[Bacry et~al.(2016)Bacry, Jaisson, and Muzy]{2014jaisson}
E.~Bacry, T.~Jaisson, and J.-F. Muzy.
\newblock Estimation of slowly decreasing hawkes kernels: application to
  high-frequency order book dynamics.
\newblock \emph{Quantitative Finance}, pages 1--23, 2016.

\bibitem[Choromanska et~al.(2015)Choromanska, Henaff, Mathieu, Arous, and
  LeCun]{choromanska2015loss}
A.~Choromanska, M.~Henaff, M.~Mathieu, G.~Ben Arous, and Y.~LeCun.
\newblock The loss surfaces of multilayer networks.
\newblock In \emph{AISTATS}, 2015.

\bibitem[Crane and Sornette(2008)]{crane2008robust}
R.~Crane and D.~Sornette.
\newblock Robust dynamic classes revealed by measuring the response function of
  a social system.
\newblock \emph{Proceedings of the National Academy of Sciences}, 105\penalty0
  (41), 2008.

\bibitem[Da~Fonseca and Zaatour(2014)]{da2014hawkes}
J.~Da~Fonseca and R.~Zaatour.
\newblock Hawkes process: Fast calibration, application to trade clustering,
  and diffusive limit.
\newblock \emph{Journal of Futures Markets}, 34\penalty0 (6):\penalty0
  548--579, 2014.

\bibitem[Daley and Vere-Jones(2003)]{daley2007introduction}
D.~J. Daley and D.~Vere-Jones.
\newblock \emph{An Introduction to the Theory of Point Processes Volume I:
  Elementary Theory and Methods}.
\newblock Springer Science \& Business Media, 2003.

\bibitem[Duchi et~al.(2011)Duchi, Hazan, and Singer]{duchi2011adaptive}
J.~Duchi, E.~Hazan, and Y.~Singer.
\newblock Adaptive subgradient methods for online learning and stochastic
  optimization.
\newblock \emph{The Journal of Machine Learning Research}, 12:\penalty0
  2121--2159, 2011.

\bibitem[Eichler et~al.(2016)Eichler, Dahlhaus, and
  Dueck]{graphicalModelingHawkes}
M.~Eichler, R.~Dahlhaus, and J.~Dueck.
\newblock Graphical modeling for multivariate hawkes processes with
  nonparametric link functions.
\newblock \emph{Journal of Time Series Analysis}, pages n/a--n/a, 2016.
\newblock ISSN 1467-9892.
\newblock \doi{10.1111/jtsa.12213}.
\newblock URL \url{http://dx.doi.org/10.1111/jtsa.12213}.
\newblock 10.1111/jtsa.12213.

\bibitem[Farajtabar et~al.(2015)Farajtabar, Wang, Rodriguez, Li, Zha, and
  Song]{gomez15}
M.~Farajtabar, Y.~Wang, M.~Rodriguez, S.~Li, H.~Zha, and L.~Song.
\newblock Coevolve: A joint point process model for information diffusion and
  network co-evolution.
\newblock In \emph{Advances in Neural Information Processing Systems}, pages
  1945--1953, 2015.

\bibitem[Gomez-Rodriguez et~al.(2013)Gomez-Rodriguez, Leskovec, and
  Sch\"olkopf]{gomez13}
M.~Gomez-Rodriguez, J.~Leskovec, and B.~Sch\"olkopf.
\newblock Modeling information propagation with survival theory.
\newblock \emph{Proceedings of the International Conference on Machine
  Learning}, 2013.

\bibitem[Granger(1969)]{granger1969}
C.~W.~J. Granger.
\newblock Investigating causal relations by econometric models and
  cross-spectral methods.
\newblock \emph{Econometrica}, 37\penalty0 (3):\penalty0 424--438, 1969.
\newblock ISSN 00129682, 14680262.
\newblock URL \url{http://www.jstor.org/stable/1912791}.

\bibitem[Hall(2005)]{hall}
A.~R. Hall.
\newblock \emph{Generalized Method of Moments}.
\newblock Oxford university press, 2005.

\bibitem[Hansen(1982)]{hansen1982large}
L.~P. Hansen.
\newblock Large sample properties of generalized method of moments estimators.
\newblock \emph{Econometrica: Journal of the Econometric Society}, pages
  1029--1054, 1982.

\bibitem[Hansen et~al.(2015)Hansen, Reynaud-Bouret, and
  Rivoirard]{hansen_reynaud_bouret_viroirard}
N.~R. Hansen, P.~Reynaud-Bouret, and V.~Rivoirard.
\newblock Lasso and probabilistic inequalities for multivariate point
  processes.
\newblock \emph{Bernoulli}, 21\penalty0 (1):\penalty0 83--143, 2015.

\bibitem[{Hardiman} and {Bouchaud}(2014)]{bouchaud14}
S.~J. {Hardiman} and J.-P. {Bouchaud}.
\newblock {Branching-ratio approximation for the self-exciting Hawkes process}.
\newblock \emph{Phys. Rev. E}, 90\penalty0 (6):\penalty0 062807, December 2014.
\newblock \doi{10.1103/PhysRevE.90.062807}.

\bibitem[Hawkes(1971)]{hawkes1971point}
A.~G. Hawkes.
\newblock Point spectra of some mutually exciting point processes.
\newblock \emph{Journal of the Royal Statistical Society. Series B
  (Methodological)}, 33\penalty0 (3):\penalty0 438--443, 1971.
\newblock ISSN 00359246.
\newblock URL \url{http://www.jstor.org/stable/2984686}.

\bibitem[Hawkes and Oakes(1974)]{hawkes1974cluster}
A.~G. Hawkes and D.~Oakes.
\newblock A cluster process representation of a self-exciting process.
\newblock \emph{Journal of Applied Probability}, pages 493--503, 1974.

\bibitem[{Jovanovi{\'c}} et~al.(2015){Jovanovi{\'c}}, {Hertz}, and
  {Rotter}]{hk_cumul}
S.~{Jovanovi{\'c}}, J.~{Hertz}, and S.~{Rotter}.
\newblock {Cumulants of Hawkes point processes}.
\newblock \emph{Phys. Rev. E}, 91\penalty0 (4):\penalty0 042802, April 2015.
\newblock \doi{10.1103/PhysRevE.91.042802}.

\bibitem[Lemonnier and Vayatis(2014)]{lemonnier2014nonparametric}
R.~Lemonnier and N.~Vayatis.
\newblock Nonparametric markovian learning of triggering kernels for mutually
  exciting and mutually inhibiting multivariate hawkes processes.
\newblock In \emph{Machine Learning and Knowledge Discovery in Databases},
  pages 161--176. Springer, 2014.

\bibitem[Lewis and Mohler(2011)]{lewis2011nonparametric}
E.~Lewis and G.~Mohler.
\newblock A nonparametric em algorithm for multiscale hawkes processes.
\newblock \emph{Journal of Nonparametric Statistics}, 2011.

\bibitem[Mohler et~al.(2011)Mohler, Short, Brantingham, Schoenberg, and
  Tita]{mohler2011self}
G.~O. Mohler, M.~B. Short, P.~J. Brantingham, F.~P. Schoenberg, and G.~E. Tita.
\newblock Self-exciting point process modeling of crime.
\newblock \emph{Journal of the American Statistical Association}, 2011.

\bibitem[Newey and McFadden(1994)]{newey1994large}
W.~K Newey and D.~McFadden.
\newblock Large sample estimation and hypothesis testing.
\newblock \emph{Handbook of econometrics}, 4:\penalty0 2111--2245, 1994.

\bibitem[Ogata(1981)]{ogata1981lewis}
Y.~Ogata.
\newblock On lewis' simulation method for point processes.
\newblock \emph{Information Theory, IEEE Transactions on}, 27\penalty0
  (1):\penalty0 23--31, 1981.

\bibitem[Ogata(1998)]{ogata1998space}
Y.~Ogata.
\newblock Space-time point-process models for earthquake occurrences.
\newblock \emph{Annals of the Institute of Statistical Mathematics},
  50\penalty0 (2):\penalty0 379--402, 1998.

\bibitem[Podosinnikova et~al.(2015)Podosinnikova, Bach, and
  Lacoste-Julien]{podosinnikova2015rethinking}
A.~Podosinnikova, F.~Bach, and S.~Lacoste-Julien.
\newblock Rethinking lda: moment matching for discrete ica.
\newblock In \emph{Advances in Neural Information Processing Systems}, pages
  514--522, 2015.

\bibitem[Reynaud-Bouret and Schbath(2010)]{reynaud2010adaptive}
P.~Reynaud-Bouret and S.~Schbath.
\newblock Adaptive estimation for hawkes processes; application to genome
  analysis.
\newblock \emph{The Annals of Statistics}, 38\penalty0 (5):\penalty0
  2781--2822, 2010.

\bibitem[Subrahmanian et~al.(2016)Subrahmanian, Azaria, Durst, Kagan, Galstyan,
  Lerman, Zhu, Ferrara, Flammini, and Menczer]{darpa16}
V.S. Subrahmanian, A.~Azaria, S.~Durst, V.~Kagan, A.~Galstyan, K.~Lerman,
  L.~Zhu, E.~Ferrara, A.~Flammini, and F.~Menczer.
\newblock The darpa twitter bot challenge.
\newblock \emph{Computer}, 49\penalty0 (6):\penalty0 38--46, 2016.

\bibitem[Xu et~al.(2016)Xu, Farajtabar, and Zha]{xu2016learning}
H.~Xu, M.~Farajtabar, and H.~Zha.
\newblock Learning granger causality for hawkes processes.
\newblock In \emph{Proceedings of The 33rd International Conference on Machine
  Learning}, pages 1717--1726, 2016.

\bibitem[Yang and Zha(2013)]{yang2013mixture}
S.-H. Yang and H.~Zha.
\newblock Mixture of mutually exciting processes for viral diffusion.
\newblock In \emph{Proceedings of the International Conference on Machine
  Learning}, 2013.

\bibitem[Zhou et~al.(2013{\natexlab{a}})Zhou, Zha, and Song]{zhou2013learning}
K.~Zhou, H.~Zha, and L.~Song.
\newblock Learning triggering kernels for multi-dimensional hawkes processes.
\newblock In \emph{Proceedings of the International Conference on Machine
  Learning}, pages 1301--1309, 2013{\natexlab{a}}.

\bibitem[Zhou et~al.(2013{\natexlab{b}})Zhou, Zha, and Song]{zhou_zha_le_2013}
K.~Zhou, H.~Zha, and L.~Song.
\newblock Learning social infectivity in sparse low-rank networks using
  multi-dimensional hawkes processes.
\newblock \emph{AISTATS}, 2013{\natexlab{b}}.

\end{thebibliography}

\bibliographystyle{plainnat}

\clearpage

\end{document}